\documentclass[acmtog]{acmart}

\usepackage{booktabs} %
\usepackage{xurl}
\usepackage{graphicx}
\usepackage{subcaption}
\usepackage[ruled]{algorithm2e} %

\SetAlFnt{\small}
\SetAlCapFnt{\small}
\SetAlCapNameFnt{\small}
\SetAlCapHSkip{0pt}

\setcopyright{acmcopyright}
\copyrightyear{2022}
\acmYear{2022}
\acmDOI{XXXXXXX.XXXXXXX}

\acmJournal{TOG}

\begin{document}

\title{CLIP-Guided StyleGAN Inversion for Text-Driven Real Image Editing}

\author{Ahmet Canberk Baykal}
\email{abaykal20@ku.edu.tr}
\orcid{0000-0002-0249-5858}
\affiliation{%
  \institution{Koç University}
  \country{Turkey}
}

\author{Abdul Basit Anees}
\email{aanees20@ku.edu.tr}
\orcid{0000-0003-1293-1796}
\affiliation{%
  \institution{Koç University}
  \country{Turkey}
}

\author{Duygu Ceylan}
\email{duygu.ceylan@gmail.com}
\affiliation{%
  \institution{Adobe Research}
  \country{United Kingdom}
}

\author{Erkut Erdem}
\email{erkut@cs.hacettepe.edu.tr}
\orcid{0000-0002-6744-8614}
\affiliation{%
  \institution{Hacettepe University}
  \country{Turkey}
}

\author{Aykut Erdem}
\email{aerdem@ku.edu.tr}
\orcid{0000-0002-6280-8422}
\affiliation{%
  \institution{Koç University}
  \country{Turkey}
}

\author{Deniz Yuret}
\email{dyuret@ku.edu.tr}
\orcid{0000-0002-7039-0046}
\affiliation{%
  \institution{Koç University}
  \country{Turkey}
}

\renewcommand{\shortauthors}{Baykal et al.}

\begin{abstract}
Researchers have recently begun exploring the use of StyleGAN-based models for real image editing. One particularly interesting application is using natural language descriptions to guide the editing process. Existing approaches for editing images using language either resort to instance-level latent code optimization or map predefined text prompts to some editing directions in the latent space. However, these approaches have inherent limitations. The former is not very efficient, while the latter often struggles to effectively handle multi-attribute changes. To address these weaknesses, we present CLIPInverter, a new text-driven image editing approach that is able to efficiently and reliably perform multi-attribute changes. The core of our method is the use of novel, lightweight text-conditioned adapter layers integrated into pretrained GAN-inversion networks. We demonstrate that by conditioning the initial inversion step on the CLIP embedding of the target description, we are able to obtain more successful edit directions. Additionally, we use a CLIP-guided refinement step to make corrections in the resulting residual latent codes, which further improves the alignment with the text prompt. Our method outperforms competing approaches in terms of manipulation accuracy and photo-realism on various domains including human faces, cats, and birds, as shown by our qualitative and quantitative results.
\end{abstract}



\begin{teaserfigure}
  \includegraphics[width=\linewidth]{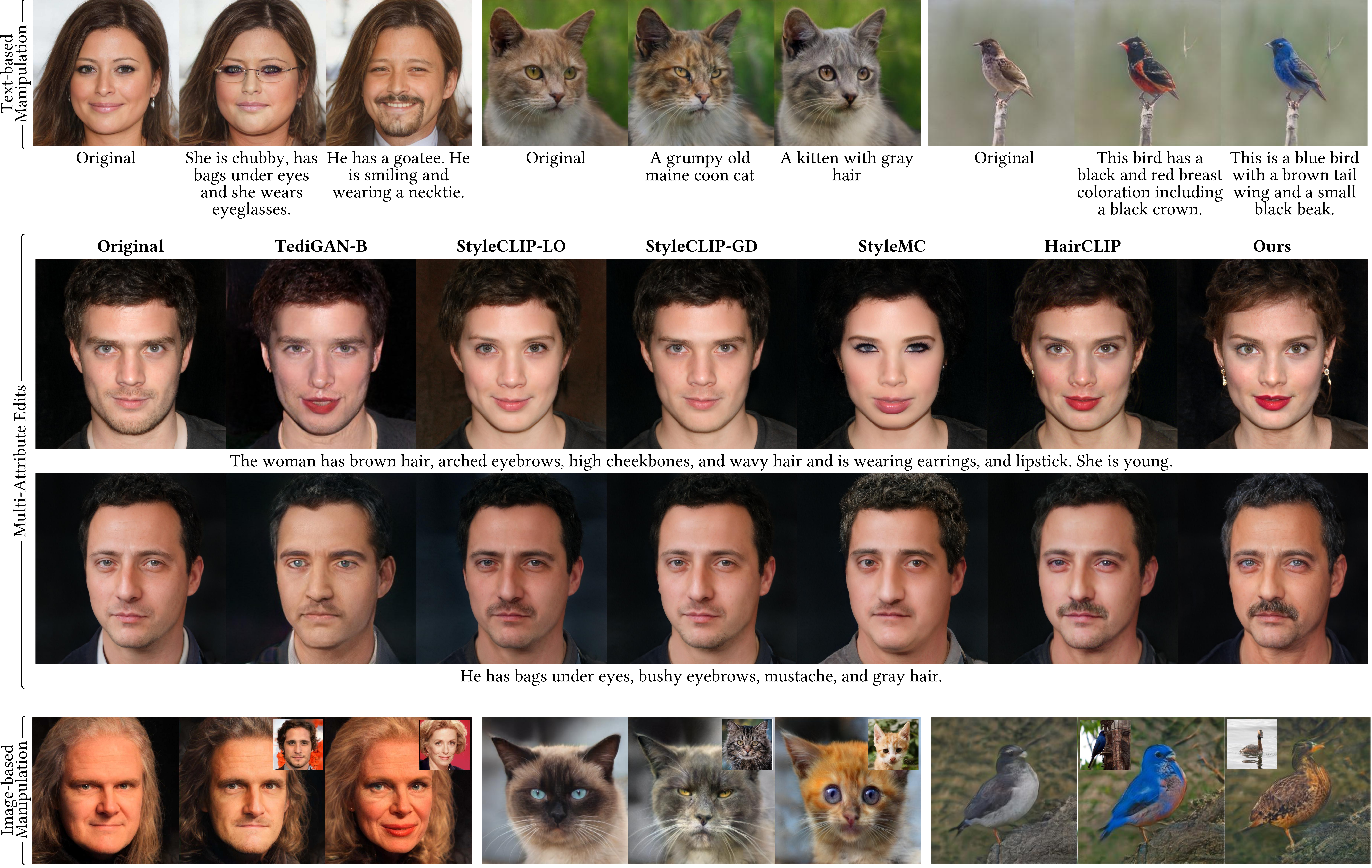}
  \caption{\textbf{Multi-attribute real image manipulation with CLIPInverter.} We present \textit{CLIPInverter} that enables users to easily perform semantic changes on images using free natural text. Our approach is not specific to a certain category of images and can be applied to many different domains (e.g., human faces, cats, birds) where a pretrained StyleGAN generator exists (\textit{top}). %
  Our approach specifically gives more accurate results for multi-attribute edits as compared to the prior work (\textit{middle}). Moreover, as we utilize CLIP's semantic embedding space, it can also perform manipulations based on reference images without any training or finetuning (\textit{bottom}). %
  }
  \label{fig:teaser}
\end{teaserfigure}

\maketitle

\section{Introduction}
\label{sec:intro}

The quality of images synthesized by Generative Adversarial Networks \cite{Goodfellow2014} have reached a remarkable level in less than a decade. StyleGAN and its variants~\cite{KarrasLA19,Karras2019stylegan2,Karras2021} are now capable of generating highly realistic images, while allowing control over the generation process by means of style mixing. Recent works~\cite{ganspace2020,shen2020interpreting} have demonstrated that StyleGAN learns disentangled attributes, making it possible to find directions in its latent space to generate images that possess such desired attributes. Consequently, there has been a growing interest in utilizing semantic editing directions in the latent space mostly for preset directions such as gender, face orientation, hair color.

Concurrent to the advances in generative modeling, we are also witnessing exciting breakthroughs in multimodal learning. For example, the recently proposed Contrastive Language-Image Pre-training (CLIP) model~\cite{radford2021learning} provides an effective common embedding for images and text captions. Such an embedding, when combined with powerful GANs paves the road towards text-guided image editing, one of the most natural and intuitive ways of manipulating images. Hence, it comes with no surprise that several recent works~\cite{li2020manigan, xia2021tedigan, Patashnik_2021_ICCV, kocasari2021, wei2022hairclip} have focused on mapping target textual descriptions to editing directions in the latent space of StyleGAN. While some methods perform optimization in the latent space guided by CLIP~\cite{xia2021tedigan, Patashnik_2021_ICCV}, others train a separate mapper network for each type of textual edit~\cite{Patashnik_2021_ICCV} or a general mapper conditioned on reference images \& textual descriptions~\cite{wei2022hairclip}. Instance-based optimization methods require long inference times. Training mappers for a single text prompt reduces the inference time to a single forward pass, but comes with the price of training time as separate mappers need to be trained for each text prompt. Moreover, these mappers that operate in the latent space do not directly consider the features of the original image as they take inverted latent codes as inputs from pretrained GAN inversion networks.

In this study, we present a new approach, which we call \emph{CLIPInverter}, to automatically edit an input image based on a target textual description containing multiple attributes by adjoining lightweight adapter modules to pretrained unconditional inversion methods. CLIPInverter includes a novel CLIP-conditioned adapter module (\emph{CLIPAdapter}) that is attached to the pretrained encoder model to map both the input image and the target textual description to a residual latent code by utilizing the common CLIP embedding space. The residual latent code is then combined with the latent code of the input image obtained by the unconditional branch of the encoder, and is fed to a CLIP-guided correction module (\emph{CLIPRemapper}) that applies a final correction by blending the latent codes with latent codes predicted from the CLIP embedding of the target textual description based on learnable blending coefficients. The final latent code is decoded by a pretrained and frozen StyleGAN2 generator to synthesize the manipulated image that reflects the desired changes while preserving the identity of the original subject as much as possible. Our encoder-adapters are lightweight networks that directly modulate image feature maps using text embeddings and they could be appended to many pretrained encoders. Our CLIP-guided correction module utilizes the CLIP text embeddings to enhance the manipulations of the generated images while preserving the photorealism. Our method does not require any additional optimization on the latents and it successfully applies manipulations using various text prompts in a single forward pass. Since we directly modulate feature maps extracted during the inversion phase, our method is capable of editing images much better than the competing approaches, especially in cases when there are multiple attributes present in the target textual description, as proven by our experiments. See Fig.~\ref{fig:architecture} for an overview of our framework.

Our method aims to strike a balance between distortion and editability~\cite{Tov2021}. Namely, our text-guided CLIPAdapter is utilized to find an editing direction that is aligned with the given target description, specific to the input image. By leveraging the inversion in the $\mathcal{W+}$ space, we aim to preserve the identity of the input image in the manipulated output, which helps in achieving relatively low distortion. However, it is important to note that complete elimination of distortion is not feasible in this process. While we are able to preserve the identity to a certain degree, we observe that not all attributes described in the target caption may be fully captured in the manipulated image. To address this, we introduce the text-guided refinement module, CLIPRemapper, which applies a final correction to the latent code, further aligning it with the desired target description. Essentially, CLIPRemapper finds a more editable region in the vicinity of the latent code we obtain from the previous stage. This process boosts the manipulation performance of our model massively, while keeping the distortion at a comparable level, as shown in our ablation study.

We demonstrate editing results for challenging cases where there are many attributes present in the target description. Our method is not restricted to a particular domain like commonly studied human faces, and we also evaluate our approach on birds and cats images. Exploring the multimodal nature of CLIP, instead of target textual descriptions, we can additionally use images or target textual descriptions containing vocabulary never seen during training as the guiding signal. Finally, we show that linearly interpolating between the original latent code and the updated latent code results in smooth image manipulations, providing a means for user to have control over the manipulation process. 

We evaluate our method on a diverse set of datasets and provide detailed qualitative results and comparisons against the state-of-the-art models. Quantitative comparisons in language-guided editing still remains a challenge, as one needs to evaluate the manipulations from different aspects, such as accuracy, preservation of text-irrelevant details, photorealism etc. Current metrics are not suitable for evaluation as they do not consider some of these aspects at all. We propose two new metrics, Attribute Manipulation Accuracy (AMA), CLIP Manipulative Precision (CMP) to measure how accurately the manipulations are applied, and how well the text-irrelevant details are preserved. We perform quantitative comparisons against state-of-the-art models using these  metrics along with FID. These comparisons as well as a user study that we conducted to evaluate perceptual realism and manipulation accuracy demonstrate the superiority of our approach over the prior work.

Our code and models are publicly available at the project website\footnote{\url{https://cyberiada.github.io/CLIPInverter}}.

\section{Related Work}
\label{sec:related}

\subsection{GAN Inversion}

In response to the growing demand for interpretability and controllability in GANs, the need for GAN inversion has emerged as a pivotal technique. By mapping a given image back into the latent space of a pretrained GAN model, as introduced by \citet{zhu2016generative}, GAN inversion facilitates a deeper understanding of the underlying features and structures in the latent space, enabling researchers to manipulate and interpret generated images with greater precision and insight. Below we discuss some representative works to highlight three main approaches to accomplish GAN Inversion -- please refer to the recent survey~\cite{xia2021survey} for an in-depth discussion of various other inversion methods.

The optimization-based methods directly optimize a latent code that reconstructs the target image as close as possible using gradient descent \cite{CreswellB16b,Abdal2019,abdal2020,Tewari2020}. This line of works is instance specific, and does not require any trainable modules. The learning-based methods invert an image by a learned encoder. This approach is similar to an autoencoder pipeline, where the pretrained generator acts as the decoder. Unconditional encoders~\cite{tewari2020pie,zhu2020indomain,alaluf2021restyle,Bau2019,richardson2021encoding,Tov2021,bai2022high} aim to solely invert the image, without any modifications while conditional encoders~\cite{alaluf2021matter} are designed for obtaining a latent code conditioned on attributes such as pose, age, or facial expressions. %
The so-called hybrid methods~\cite{zhu2016generative,Bau2019Large} combine optimization-based methods with learning-based methods. The images are first inverted to a latent code by a learned encoder. This latent code then becomes the initialization for the latent optimization, and is optimized to reconstruct the target image. 

More recent approaches build different architectures, fine-tune StyleGAN weights, or modulate feature maps for inversion. Style~Transformer~\cite{hu2022style} uses a combination of convolutional neural networks and transformers to invert images into the latent space. Pivotal Tuning Inversion (PTI)~\cite{roich2021pivotal} fine-tunes the generator around a pivotal latent code to find a balance for the distortion-editability tradeoff. Some methods~\cite{alaluf2021hyperstyle, dinh2021hyperinverter} train hypernetworks to modulate the weights of a pre-trained StyleGAN network for accurate as well as editable inversions. Spatially-Adaptive Multilayer (SAM) GAN Inversion~\cite{parmar2022sam} predicts invertibility maps and High-Fidelity GAN Inversion (HFGI)~\cite{wang2021HFGI} predicts latent maps to modulate StyleGAN features.  %

While both optimization-based and hybrid approaches may reconstruct images faithfully, they require solving an optimization problem for each image, resulting in longer processing times. On the other hand, our approach adapts learned adapters appended to encoders, which provides a much faster alternative to current methods. Furthermore, we condition the inversion process directly on the target captions, which ensures that a more effective editing space direction can be found in the latent space.

\begin{figure*}[!t]
  \centering
  \includegraphics[width=0.99\linewidth]{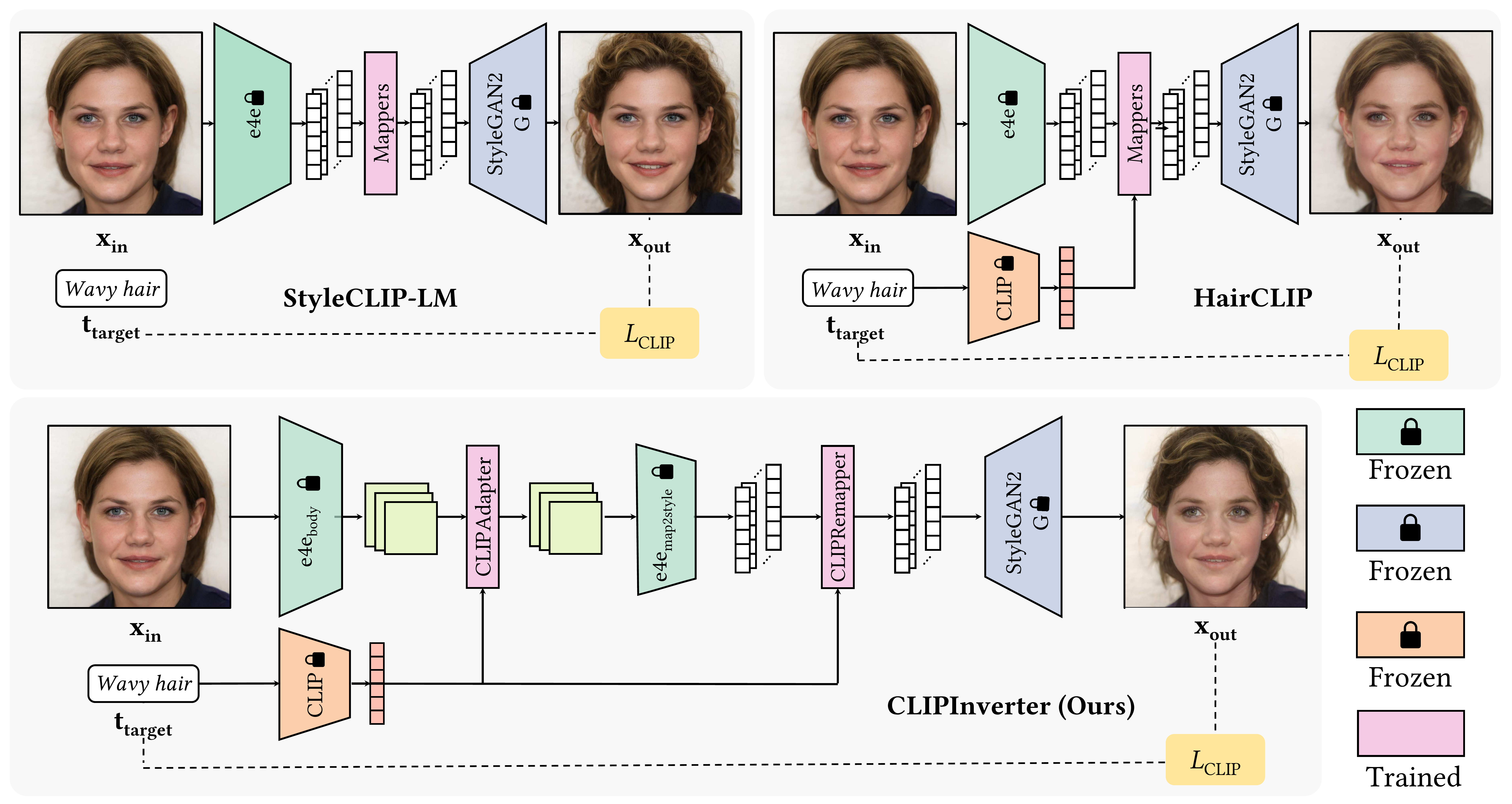}
  \caption{\textbf{An overview of our CLIPInverter approach in comparison to similar text-guided image manipulation methods.} StyleCLIP-LM utilizes target description only in the loss function. HairCLIP additionally uses the description to modulate the latent code obtained by the encoder within the mapper. Alternatively, our CLIPInverter employs specially designed adapter layers, CLIPAdapter, to modulate the encoder in extracting the latent code with respect to the target description. To further obtain more accurate edits, it also makes use of an extra refinement module, CLIPRemapper, to make subsequent corrections on the predicted latent code.} 
  \label{fig:architecture}
\end{figure*}

\subsection{Latent Space Manipulation}
Recent work has shown that GANs learn a semantically-coherent latent space, enabling to map manipulations in the latent space to semantic image editing. Specifically, StyleGAN~\cite{KarrasLA19} learns an intermediate latent space by employing a mapping network to transform the sampled latent code. These intermediate latent codes determine the parameters of the AdaIN~\cite{huang2017adain} layers introduced in the generator to control the style of the generated image, allowing control over the synthesis at different levels. %
A common approach when manipulating images is to first invert the input image back into the latent space of a pretrained generator using GAN inversion and then traverse the latent space to find a meaningful direction. Such a direction can be found by either using explicit supervision of image attribute annotations~\cite{shen2020interpreting,Abdal2021, wu2020stylespace}, or in an unsupervised manner~\cite{voynov2020unsupervised, ganspace2020,shen2021closedform}. %
Recently proposed methods consider various modalities for conditional image manipulation. StyleMapGAN~\cite{kim2021stylemapgan} proposes an intermediate latent space with spatial dimensions with spatial modulation that enables local editing based on reference images. Similarly, the study by ~\cite{collins2020editing} uses a transformation matrix to control the interpolation between an input image and a reference image in the latent space to locally edit the input image. The recent work of~\cite{alaluf2021matter} manipulates an input image based on a target age by training an encoder conditioned on the target age to find residual latent codes to add to the inverted latent code of the original image. In a similar vein, we train adapter layers appended to an encoder conditioned on textual descriptions to output these residual latent codes. We also use the CLIP model to define supervisory signals to explore the similarity of an input image and a textual description.

Moreover, there are several latent spaces to consider in a StyleGAN2 generator. The latent mapper transforms the latent codes in the space $\mathcal{Z}$ drawn from a Normal distribution to an intermediate latent space $\mathcal{W}$. The latent codes in the $\mathcal{W}$ space are used at different stages in the StyleGAN2 generator, after being mapped to the $\mathcal{S}$ space by an affine transformation. $\mathcal{W}+$ space is an extended version of the $\mathcal{W}$ space where a different $\mathbf{w}$ is used for each style input of the generator. While some works find editing directions in the $\mathcal{S}$ space such as StyleCLIP-GD \cite{Patashnik_2021_ICCV} and StyleMC \cite{kocasari2021}, many others like StyleCLIP-LO, StyleCLIP-LM \cite{Patashnik_2021_ICCV}, SAM \cite{alaluf2021matter} utilize the extended intermediate space $\mathcal{W}+$. Our text-guided image encoder operates on $\mathcal{W}+$ to find effective editing directions.

\subsection{Text-Guided Image Manipulation}
Given an image and a target description in natural language, the aim of text-guided image manipulation is to generate images that reflect the desired semantic changes while also preserving the details or attributes not mentioned in the text. ManiGAN \cite{li2020manigan} learns a text-image affine combination   which selects image regions that are relevant to the language description and a detail correction module that modifies these regions. TediGAN \cite{xia2021tedigan} enforces the text and image matching by mapping the images and the text to the same latent space and performs further optimization to preserve the identity of the subjects in the original image. %

More recent works use semantics learned by a multi-modal method such as CLIP \cite{radford2021learning}. %
StyleCLIP \cite{Patashnik_2021_ICCV} uses the CLIP space to optimize for the latent code (StyleCLIP-LO) that minimizes the distance of the image and text pair. They also present a latent mapper (StyleCLIP-LM) that predicts residual latent codes corresponding to specific attributes. Finally, they also experiment with mapping a text prompt to a global direction (StyleCLIP-GD) in the latent space that is independent of the input image. The most recent StyleMC \cite{kocasari2021} model presents an efficient method to learn global directions in the $\mathcal{S}$ space of StyleGAN2 for a given text prompt, by finding directions at lower resolutions and applying manipulations at higher resolutions. It also utilizes CLIP to minimize the distance between the generated image and the text prompt. Most recently and most similar to our approach, HairCLIP~\cite{wei2022hairclip} modulates the inverted latent codes based on hairstyle and hair color inputs as image or text. Their approach is similar to StyleCLIP-LM. However, they also modulate the latent codes with the CLIP embeddings rather than solely optimizing the similarity in the CLIP space.

Our work share some similarities with the aforementioned methods. Like the original TediGAN model, we employ an encoder to predict the latent code conditioned on the provided target description. That said, we estimate a residual latent code reflecting only the desired changes mentioned in the description, which is to be added to the inverted latent code of the input image. StyleCLIP-LM and StyleMC models predict residual latent codes similar to ours, but they require training their mapper functions from scratch for each text prompt via a loss function based on CLIP similarity. Most similar to our approach, HairCLIP applies modulations in the latent space after obtaining inversions with a pretrained network. On the other hand, we let CLIP embeddings modulate the feature maps via an adapter module for predicting the residual latent code. With this modulation, our inversion step is text-guided, whereas HairCLIP applies text-conditioning on the latent space. We also train a correction module which applies latent code blending with learnable blending coefficients for improved accuracy, quality and fidelity in the output images. In Fig.~\ref{fig:architecture}, we illustrate the aforementioned fundamental differences between our approach and the most similar StyleCLIP-LM and HairCLIP methods.

Our approach allows us to manipulate fine-scale details by modulating the feature maps, resulting in more accurate manipulations than HairCLIP. Thanks to this process, we also eliminate the need for separate training, unlike StyleCLIP-LM. That is, once our model is trained, it can be directly used to manipulate images by considering a large variety of text prompts containing multiple attributes.
We provide extensive comparisons against the aforementioned recent StyleGAN-based methods in Section~\ref{sec:experiments} and show the superiority or competitiveness of our proposed approach.

Recently, diffusion models trained with variational inference achieved state-of-the-art performance in image generation~\cite{dhariwal2021diffusion, ho2020denoising, Rombach_2022_CVPR}. With this success of diffusion based models, several text-guided image manipulation methods have been proposed. DiffusionCLIP~\cite{Kim_2022_CVPR} first converts the images to latent noises by forward diffusion and then guides the reverse diffusion process by CLIP to control the attributes in the synthesized images. UniTune~\cite{Valevski2022UniTuneTI} introduces a simple method to fine-tune large scale text-to-image diffusion models on single images. Similarly, Imagic~\cite{kawar2022imagic} optimizes a text embedding and fine-tunes pretrained generative diffusion models to perform edits on a single image. Prompt-to-Prompt~\cite{hertz2022prompttoprompt} and its later extension Plug-and-Play~\cite{pnpDiffusion2022} achieve semantic edits by blending activations extracted from both the original and target prompts. These diffusion-based editing methods differ from ours as each one requires a large pre-trained text-to-image network. Hence, we do not directly evaluate our approach against these methods, but provide some comparisons in the supplementary.

\subsection{Adapter Layers}
Adapter layers~\cite{adapter}, originally proposed for NLP tasks, are compact modules that allow parameter sharing in an efficient manner. The key idea is to add adapter modules, consisting of a few layers, between the layers of a pretrained network. The parameters of the adapter module are updated during the fine-tuning phase on a downstream task, while the original parameters of the pretrained network remain the same. This way, most of the parameters of the pretrained network are shared between different downstream tasks, resulting in a model that is able to perform diverse tasks efficiently. Since the parameters of the pretrained network are frozen, the original capabilities of the model are preserved. The module proposed for NLP~\cite{adapter} is appended after the feed-forward layers and before adding the skip connection back, in a transformer model. This module consists of a down-projection and an up-projection layer. Compared to the original pretrained model, the number of parameters of the adapter module is considerably smaller, allowing the learning of new tasks efficiently.

Adapter layers have also been proposed to use in computer vision tasks. ~\citet{RebuffiBV17} introduced residual adapter layers for multiple-domain learning in image recognition. Their residual adapter layers are slightly modified versions of the residual blocks in ResNet~\cite{resnet}, where batch normalization and 1$\times$1 convolutions with residual connections are added to these residual blocks. ~\cite{Rebuffi2} proposed several improvements over this module. They modified the series implementation of the residual adapter to obtain a parallel adapter, where the input to the convolutional blocks of the residual block is processed in parallel with the adapter convolutions and fed back to the original branch. They also investigated where to place the adapter layers in the ResNet to achieve the best performance. Finally, in VL-Adapter~\cite{sung2022vladapter}, the authors experimented with adapter layers in vision and language joint tasks. They added adapter modules consisting of downsampling and upsampling layers to the transformer architecture for parameter efficient fine-tuning. %

Our approach consists of adapter modules that we attach to inversion models. Our encoder adapter module is similar to the mapping networks in StyleCLIP-LM. However, in these adapter modules, we modulate intermediate image feature maps that are extracted from the inversion model. After the modulation, the feature maps are fed back to the inversion model to be processed further. With this essential idea, we are able to add a text conditional branch to the existing GAN inversion models while preserving its unconditional inversion capabilities.
\section{The Approach}
\label{sec:approach}
\subsection{Overview of CLIPInverter}
Our text-guided image editing framework includes two separate modules, namely CLIPAdapter and CLIP Remapper, each playing a different role in obtaining the desired edit. CLIPAdapter involves CLIP-conditioned adapter layers for the GAN inversion process, which are used for finding semantic editing directions in the latent space along which the given input image is manipulated. CLIPRemapper then performs a final refinement over the predicted latent code of the output image considering the CLIP embedding of the input text prompt to further improve the manipulation accuracy as well as the perceptual quality.

\begin{figure*}[!t]
  \centering
  \begin{tabular}{p{0.455\linewidth}p{0.515\linewidth}}
       \includegraphics[width=0.805\linewidth]{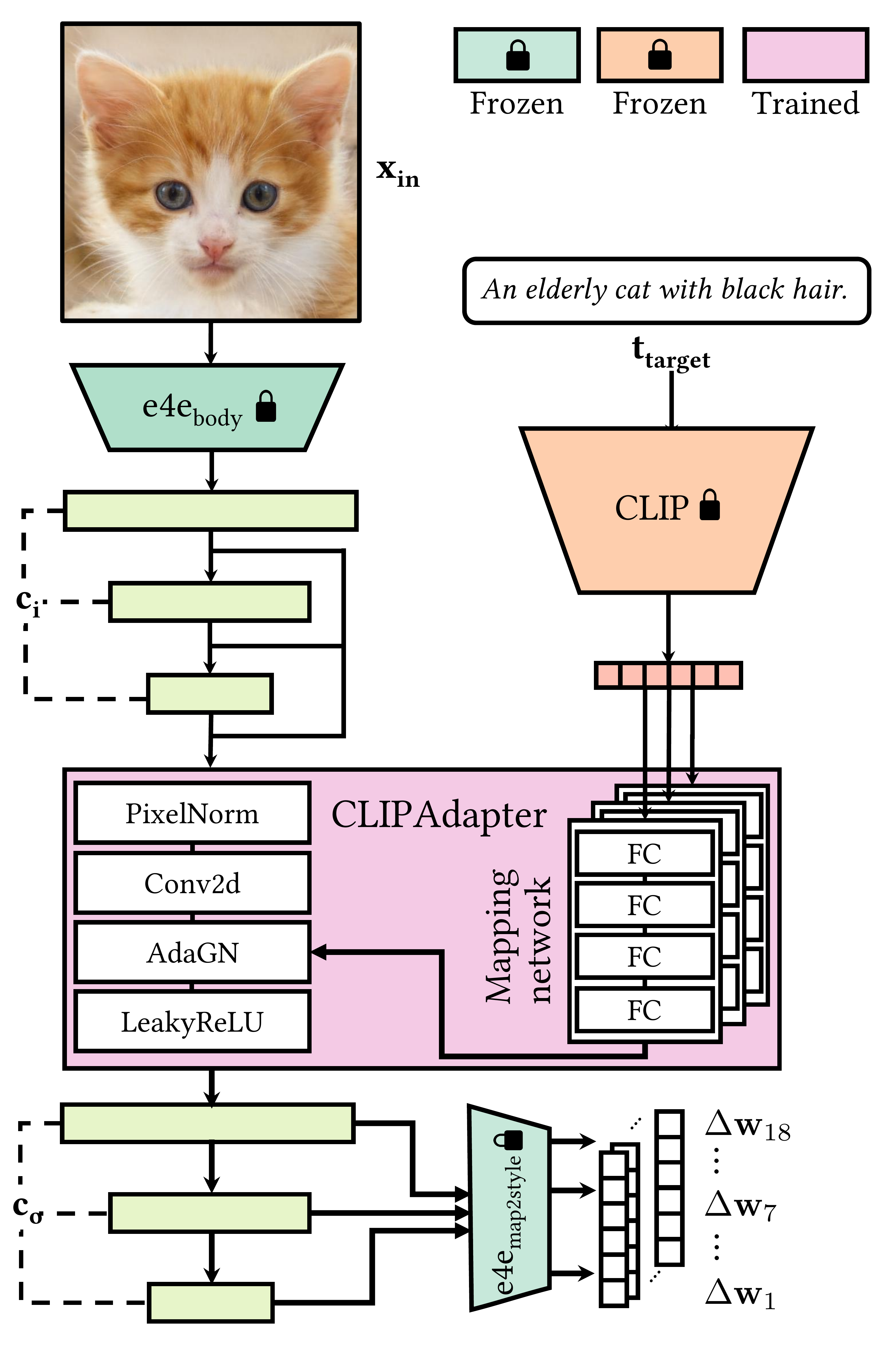} & \includegraphics[width=\linewidth]{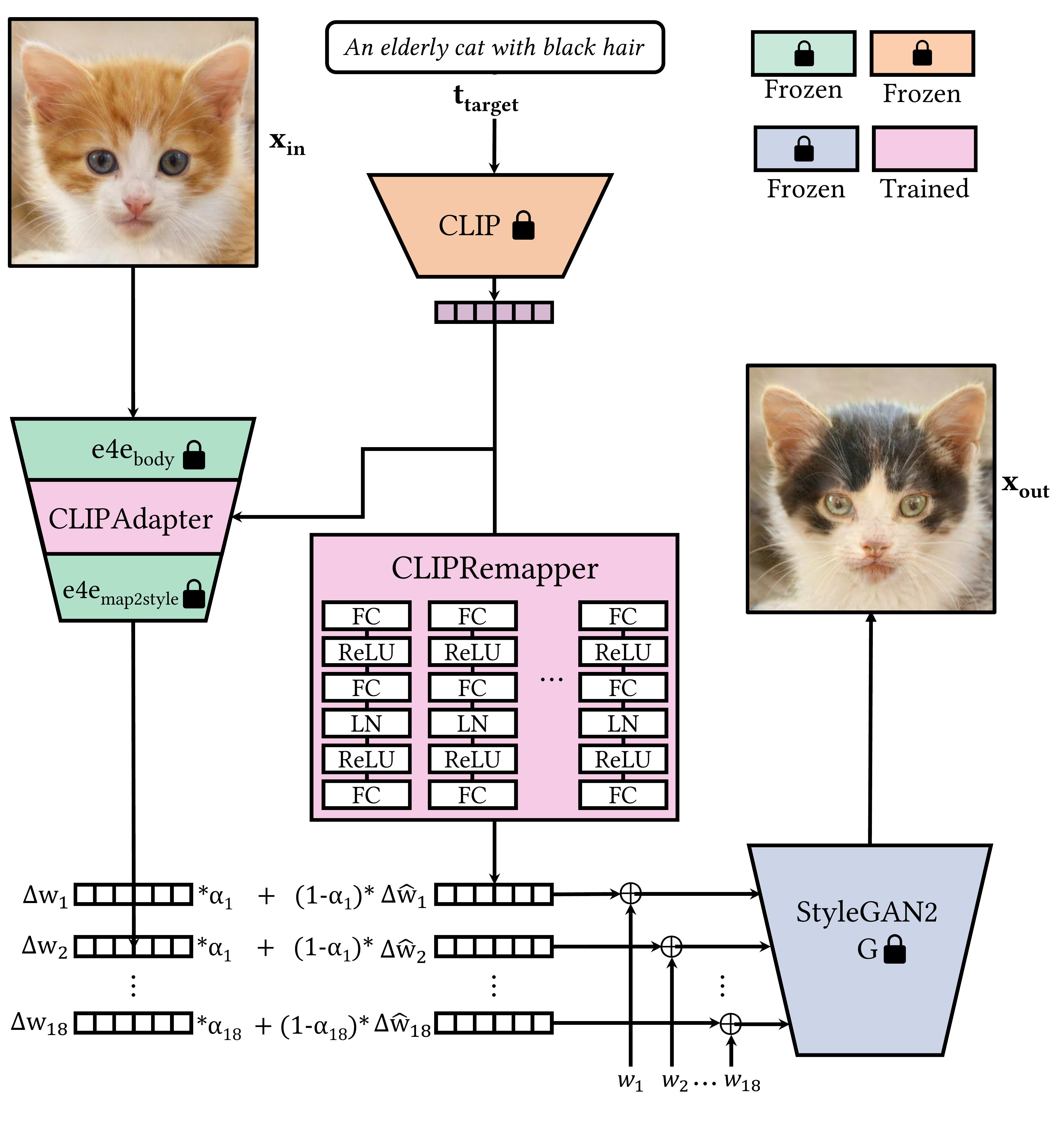} \\
       {\small (a) \textbf{Architectural details of CLIPAdapter}. Our text-guided image encoder network inherits the structure of e4e, and makes it conditional on CLIP embedding of the target text. This is achieved by shallow mapping networks at three different scales to better align the multi-modal semantic space of CLIP model with the $\mathcal{W}$+ space of StyleGAN2, whose outputs control the prediction of residual codes through AdaGN layers.} & 
       {\small (b) \textbf{Architectural details of CLIPRemapper}. Our refinement module consists of MLPs that predict a residual latent code solely based on the CLIP embedding of the target description. This residual is blended with the residual predicted by the CLIPAdapter module, providing a corrected (better aligned) residual latent code which is used to synthesize the final manipulated image via the pre-trained StyleGAN2 generator.}
  \end{tabular}\vspace{-0.1cm}
  \caption{\textbf{CLIPAdapter and CLIPRemapper modules of our CLIPInverter framework.} Our text-guided image editing framework includes two key modules, CLIPAdapter and CLIP Remapper.  CLIPAdapter employs CLIP-conditioned adapter layers within the GAN inversion process to find the semantic editing direction in the latent space. CLIPRemapper further refines the predicted edit direction to improve the manipulation accuracy again based on the CLIP embedding of the input text prompt.}  
  \label{fig:CLIPInverter}
\end{figure*}

Given an input image $\mathbf{x}_{\mathbf{in}}$ and a desired target description $\mathbf{t}_{\mathbf{target}}$, the goal of our CLIPInverter approach is to manipulate the input image and synthesize an output image $\mathbf{x}_{\mathbf{out}}$ such that the end result reflects the attributes described in the text (e.g., hair color, age, gender), while preserving the identity of the subject present in the original image or any other features not relevant to the description. Assuming that we have access to a StyleGAN2~\cite{Karras2019stylegan2} generator $G$ that can synthesize images from a particular domain, we cast this text-guided manipulation task as finding a mapping of the input image $\mathbf{x}_{\mathbf{in}}$ and the target text prompt $\mathbf{t}_{\mathbf{target}}$ to a latent code $\mathbf{w^*}\in\mathcal{W}+$ in the latent space of $G$ so that when decoded it generates the manipulation result as $\mathbf{x}_{\mathbf{out}} = G(\mathbf{w^*})$. 

We perform the latent space mapping in two steps, using the unconditional and the conditional branches of the text-guided encoder, which we obtain by attaching CLIPAdapter to a pretrained image inversion network, namely \mbox{encoder4editing}~($\text{e4e}$)~\cite{Tov2021}.  %
We first map the input image $\mathbf{x}_{\mathbf{in}}$ to its latent code $\mathbf{w}$ through the pretrained encoder $\text{e4e}$. We then compute a residual latent vector $\Delta \mathbf{w}$ through the conditional branch, which processes both the input image and the CLIP model~\cite{radford2021learning} embedding of the textual description. The final image $\mathbf{x}_{\mathbf{out}}$ is synthesized by passing the aggregated latent code first through the refinement module, $\mathbf{w^*} = \text{CLIPRemapper}(\mathbf{w} + \Delta \mathbf{w})$, then through the generator network, which is a pretrained StyleGAN2~\cite{Karras2019stylegan2} generator. %
CLIPInverter applies one final correction to the latent code by predicting latents based on the CLIP embedding of the target caption $\mathbf{t}_{\mathbf{target}}$. Then, the predicted latent is blended with the previously inverted latent code depending on a learnt interpolation coefficient $\alpha$. 

In the following, we describe the details of the key modules of CLIPInverter and the loss functions we utilize during training.

\subsection{CLIPAdapter: CLIP-Guided Adapters for Latent Space Manipulation}
Fig.~\ref{fig:CLIPInverter} (a) shows the architecture of our proposed text-guided encoder, which follows the architecture of $\text{e4e}$ with attached lightweight adapters that enable us to incorporate the textual descriptions. The original $\text{e4e}$ architecture maps the input image to feature maps at three levels -- coarse, medium and fine. We introduce Adaptive Group Normalization (AdaGN) layers in CLIPAdapter, replacing the Instance Normalization in the AdaIN~\cite{huang2017adain} layers to modulate these features using features obtained from the CLIP~\cite{radford2021learning} embedding of the target description.

CLIPAdapter also employs shallow mapping networks, one for each level, to better align the multi-modal semantic space of the CLIP model with the $\mathcal{W}+$ space of StyleGAN2. Specifically, we feed the text embedding obtained from the CLIP model to a multi-layer perceptron (MLP) which predicts the scale and shift parameters of the subsequent AdaGN blocks. Given the image features from the coarse, medium, and fine layers of the encoder, the AdaGN blocks perform feature modulation such that the outputs control the prediction of the residual latent codes. 

The design philosophy behind our encoder architecture is to have adapter layers in a pretrained network that can identify visual features relevant and irrelevant to the manipulation task in both image and text-specific manner in computing the residual latent code to identify the manipulation direction in the $\mathcal{W+}$ space. Specifically, we factorize the layers of the $\text{e4e}$ network into two groups: $\text{e4e}_{\text{body}}$ and $\text{e4e}_{\text{m2s}}$. While $\text{e4e}_{\text{body}}$ includes the convolutional backbone layers and it extracts a feature pyramid consisting of feature maps from coarse, medium and fine levels, $\text{e4e}_{\text{body}}$ consists of small convolutional mapping networks that transforms these feature maps to the latent styles in the $\mathcal{W+}$ space. We insert CLIPAdapter between $\text{e4e}_{\text{body}}$ and $\text{e4e}_{\text{m2s}}$.

More formally, in order to manipulate a given image $\mathbf{x}_{\mathbf{in}}$ based on a text prompt $\mathbf{t}_{\mathbf{target}}$, we start with obtaining the latent code $\mathbf{w}$ of the original image in the $\mathcal{W}+$ latent space of StyleGAN2~\cite{Karras2019stylegan2} via $\text{e4e}$:
\begin{equation}
    \mathbf{w} = \text{e4e}(\mathbf{x}_{\mathbf{in}}) \in \mathbb{R}^{18 \times 512}.
    \label{eq:firstencoder}
\end{equation}

To perform semantic edits on $\mathbf{x}_{\mathbf{in}}$ to reflect the desired target look, we utilize the text-conditioned branch of our encoder network, which takes both the input image and the target textual description as input and outputs the residual latent code. During this process, we first extract intermediate feature maps $\mathbf{c_i}$ from the body layers of the encoder network, $\text{e4e}_{\text{body}}$:
\begin{equation}
    \mathbf{c_i} = \text{e4e}_{\text{body}}(\mathbf{x}_{\mathbf{in}}).
    \label{eq:nonmodulated_features}
\end{equation}

Next, we utilize the CLIP text embedding of the target text prompt $\mathbf{t}_{\mathbf{target}}$ to modulate $\mathbf{c_i}$, obtaining the modulated feature maps $\mathbf{c_o}$ through our encoder-adapter layers CLIPAdapter:
\begin{equation}
    \mathbf{c_o} = \text{CLIPAdapter}(\mathbf{c_i}, \mathbf{t}_{\mathbf{target}}).
    \label{eq:modulation}
\end{equation}

As the final step to predict the manipulation directions as residual latents $\Delta \mathbf{w}$, we pass the modulated feature maps $\mathbf{c_o}$ through the map2style layers of $
\text{e4e}$, $\text{e4e}_{\text{m2s}}$:
\begin{equation}
    \Delta \mathbf{w} = \text{e4e}_{\text{m2s}}(\mathbf{c_o}) \in \mathbb{R}^{18 \times 512}.
    \label{eq:map2style}
\end{equation}

Note that the body and map2style layers of $\text{e4e}$ are combined to complete the pretrained encoder $\text{e4e} = [\text{e4e}_\text{body}, \text{e4e}_\text{m2s}]$. The language conditioning happens in the adapter layers CLIPAdapter and these layers are the only layers with trained parameters in the inversion framework, the rest of the parameters are pretrained.

\subsection{CLIPRemapper: CLIP-Guided Latent Vector Refinement}

To further enhance the quality of the manipulated image, we introduce a final refinement step over the predicted latent code. As shown in Fig.~\ref{fig:CLIPInverter}(b), our CLIPRemapper carries out this refinement process by mapping CLIP text embedding of the given text prompt to the $\mathcal{W+}$ space and then using the projected text embedding to steer the residual latent code predicted by CLIPInverter towards a direction more compatible with the target text. Specifically, CLIPRemapper involves shallow mapping networks for each level to better align image with the text. The text embedding obtained from CLIP is fed to MLPs at each stage to predict a component for latent code correction corresponding to the caption, as follows: 
\begin{equation}
    \mathbf{\Delta \widehat{w}_i} = \text{MLP}_{i}(\mathbf{t}_\mathbf{target}).
    \label{eq:generator-adapter}
\end{equation}

Taking into account $\Delta \mathbf{\widehat{w}_i}$, we apply a further correction to the residual latent code predicted through CLIPAdapter %
as:
\begin{equation}
    \mathbf{\Delta {w_i}'} = \frac{\left(\alpha_{i} * \Delta \mathbf{{w_i}} + (1 - \alpha_{i}) * \Delta \mathbf{\widehat{w}_i}\right)*\|\Delta \mathbf{{w_i}}\|}{\|{\alpha_{i} * \Delta \mathbf{{w_i}} + (1 - \alpha_{i}) * \Delta \mathbf{\widehat{w}_i}}\|}
\end{equation}
where $\alpha_i$ is a weighting factor which is defined as a learnable parameter, and $\Delta \mathbf{{w_i}'}$ represents the final corrected residual latent code. %

In particular, the corrected residual latent code $\Delta \mathbf{{w_i}'}$ is obtained by considering linear combination of two separate codes, the residual latent code from CLIPAdapter $\Delta \mathbf{{w_i}}$ and the vector $\Delta \mathbf{\widehat{w}_i}$, followed by a normalization. We do not want the refinement procedure to make substantial changes in the predicted latent code. Hence, along with the loss functions introduced in the next section, the normalization further enforces the final latent code $\Delta \mathbf{{w_i}'}$ to be in the vicinity of the residual latent code predicted in the previous step. We only make the necessary changes in the semantic directions suggested by the CLIP embedding of the target text $\mathbf{t}_\mathbf{target}$ through a simple image composition process in the latent StyleGAN space. 

\begin{figure}[!b]
  \centering
  \includegraphics[width=\linewidth]{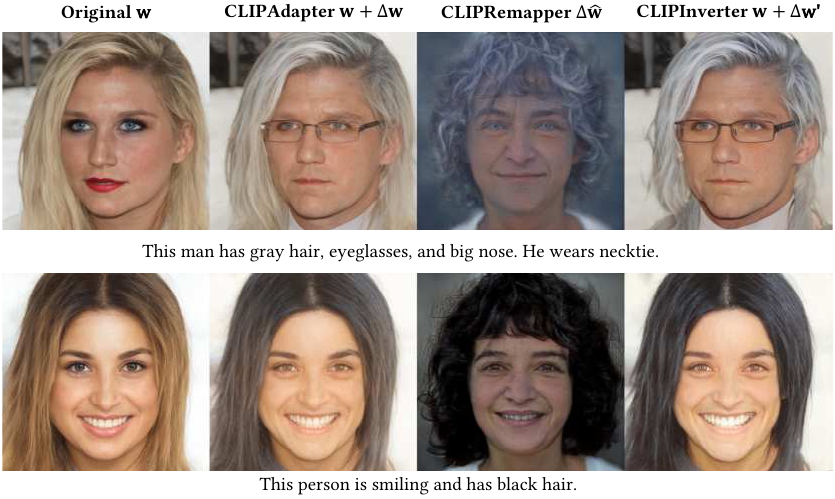}\vspace{-0.3cm}
  \caption{\textbf{Visualization of the latent code correction operation via CLIPRemapper.} For two sample images, we show the initial editing results generated solely by CLIPAdapter, the generic images generated via CLIPRemapper, and the final manipulations by CLIPInverter obtained by the suggested correction scheme. Our refinement module works as intended, providing edits more consistent with the target descriptions.} 
  \label{fig:image-components}
\end{figure}

CLIPRemapper effectively integrates the local inductive bias of the target description and the desired visual characteristics for the source image as suggested by the target description. In structured domains such as human faces,  residual latent code $\Delta \widehat{w}$ obtained in an image blind manner using target description produces interpretable results. This process, as demonstrated in Fig~\ref{fig:image-components}, combines the manipulated image generated by CLIPAdapter with a generic image that predominantly exhibits the characteristics mentioned in the target description, leading to further improvements on both the manipulation accuracy and the perceptual quality. In the case of less structured domains, e.g. birds, while $\Delta \widehat{w}$ may not be interpretable, it still provides some improvements to the manipulations. Additional visualizations for cat and bird images can be found in the supplementary material.

\subsection{Training Losses}
We train our proposed $\text{CLIPInverter}$ model on a training set of images paired with their corresponding textual descriptions $\{(\mathbf{x}_\mathbf{in},\mathbf{t}_\mathbf{real})\}$. Specifically, we employ a cyclic adversarial training  strategy~\cite{Zhu_2017cyclegan} during training, which involves two separate manipulation steps. In the first one, we feed in the original input image $\mathbf{x}_\mathbf{in}$ along with a target textual description $\mathbf{t}_{target}$ (which does not match with the input image) to our model. This process generates a manipulated image $\mathbf{x}_\mathbf{out} = \text{CLIPInverter}(\mathbf{x}_\mathbf{in},\mathbf{t}_\mathbf{target})$. In the cyclic pass, we take this manipulated image $\mathbf{x}_\mathbf{out}$ and the original text description $\mathbf{t}_\mathbf{real}$ (which describes the original input image $\mathbf{x}_\mathbf{in}$) as inputs to obtain $\widehat{\mathbf{x}}_\mathbf{in} = \text{CLIPInverter}(\mathbf{x}_\mathbf{out},\mathbf{t}_\mathbf{real})$. We expect $\widehat{\mathbf{x}}_\mathbf{in}$ to closely resemble the original image $\mathbf{x}_\mathbf{in}$ by enforcing cycle consistency. We obtain the target text description by rolling the minibatch, meaning that each image will be paired with the textual description that describes the next image in the minibatch.
We train our model with a set of loss functions. Each of these objectives are used both in the first manipulation pass and the following cycle pass. In the following, we only describe the losses for the first manipulation pass for the sake of presentation simplicity. %

We use $\mathcal{L}_{2}$ and $\mathcal{L}_{\mathrm{LPIPS}}$ \cite{DBLP:conf/cvpr/ZhangIESW18} losses  to respectively enforce pixel-wise and perceptual similarities between the input and the manipulated image, such that:
\begin{equation}
    \mathcal{L}_{\mathrm{2}} = \|\mathbf{x}_{in}-\mathbf{x}_{out}\|_{2}\,,
    \label{eq:l2loss}
\end{equation}
\begin{equation}
    \mathcal{L}_{\mathrm{LPIPS}} = \|F(\mathbf{x}_{in})-F(\mathbf{x}_{out})\|_{2}\,,
    \label{eq:lpipsloss}
\end{equation}
where $F(\cdot)$ denotes deep features extracted from a pretrained AlexNet \cite{NIPS2012_c399862d} model.

Ideally, we want any manipulation to preserve the identity of the subject in the original image. To preserve the identity, we employ an identity loss which maximizes the cosine similarity between the input image and the output image feature embeddings:
\begin{equation}
    \mathcal{L}_{\mathrm{ID}}=1-\langle R(\mathbf{x}_{in})), R(\mathbf{x}_{out})\rangle\,,
    \label{eq:idloss}
\end{equation}
where $\langle\cdot, \cdot\rangle$ represents the cosine similarity between the feature vectors, $R$ denotes a pretrained deep network. Specifically, we use the pretrained ArcFace \cite{DBLP:conf/cvpr/DengGXZ19} network for human faces, and a ResNet50 \cite{he2015deep} network trained with MOCOv2 \cite{chen2020improved} for birds and cats.

We also employ the following regularization loss, which enforces the predicted latent codes to be close to the average latent code of the generator, and shown to improve overall image quality in previous work~\cite{richardson2021encoding}, such that:
\begin{equation}
    \mathcal{L}_{\text {reg}}=\|
    \mathbf{w^*}-\overline{\mathbf{w}}\|_{2}\,,
    \label{eq:regloss}
\end{equation}
where $\mathbf{w^*}$ and $\overline{\mathbf{w}}$ are the aggregated and the average latent codes, respectively. 

Lastly, to enforce the similarity between the output image and the target description, we employ a directional CLIP loss \cite{stylegan-nada}. Rather than directly minimizing the distance between the generated image $\mathbf{x}_{out}$ and the text prompt $\mathbf{t}_{target}$ in the CLIP space, directional CLIP loss aligns the direction from the input image $\mathbf{x}_{in}$ to the manipulated image $\mathbf{x}_{out}$ with the direction from the original text description $\mathbf{t}_{real}$ to the target text description $\mathbf{t}_{target}$:
\begin{eqnarray}
\Delta T=E_\mathrm{CLIP,T}\left(\mathbf{t}_{target}\right)-E_\mathrm{CLIP,T}\left(\mathbf{t}_{real}\right), \nonumber\\
\Delta I=E_\mathrm{CLIP,I}\left(\mathbf{x}_{out}\right)-E_\mathrm{CLIP,I}\left(\mathbf{x}_{in}\right), \nonumber\\
\mathcal{L}_{\text {direction }}=1-\frac{\Delta I \cdot \Delta T}{|\Delta I||\Delta T|},
\end{eqnarray}
where $E_{\mathrm{CLIP,T}}$ and $E_{\mathrm{CLIP,I}}$ are the text and image encoders of CLIP, respectively.

Our final loss function for the first manipulation pass is a weighted sum of the objectives:
\begin{multline}
    \mathcal{L}_{\mathrm{manipulation}} = \lambda_{1}\mathcal{L}_{2} + \lambda_{2}\mathcal{L}_{\mathrm{LPIPS}} + \lambda_{3}\mathcal{L}_{\mathrm{ID}} + \lambda_{4}\mathcal{L}_{\mathrm{reg}} + \lambda_{5}\mathcal{L}_{\mathrm{direction}} \,,
    \label{eq:forwardloss}
\end{multline}
where each $\lambda_{i}$ determines the weight of the corresponding objective. The total loss including the first manipulation and the follow-up cycle passes is the following:
\begin{equation}
    \mathcal{L}_{\mathrm{total}} = \mathcal{L}_{\mathrm{manipulation}} + \lambda_6 \mathcal{L}_{\mathrm{cyclic}} \,,
    \label{eq:totalloss}
\end{equation}
where $\mathcal{L}_{\mathrm{cyclic}}$ is the cyclic consistency loss, which contains the same loss terms as $\mathcal{L}_{\mathrm{manipulation}}$ in which $\mathbf{x}_{out}$ is replaced with $\widehat{\mathbf{x}}_{in}$, and $\lambda_{\mathrm{cycle}}$ is the weight for this cyclic loss. 

During training, we follow a multi-stage regime. We first train the CLIPAdapter (without using CLIPRemapper). Once these are fully trained, we freeze the weights of CLIPAdapter weights and train the CLIPRemapper while optimizing for the CLIP loss along with the L2, LPIPS and ID losses. For the LPIPS and L2 losses, we also include the loss between images generated with and without CLIPRemapper which ensures that the CLIPRemapper does not change the images by a large amount. In addition, we also include a L2 regularization loss on the interpolation coefficients (lambdas) such that the amount of interpolation between two latent codes does not change the original code by a large amount. This is also observed to remove artifacts in the generated images.
\section{Experimental Evaluation}
\label{sec:experiments}

\subsection{Datasets}
We conduct extensive evaluation on a variety of domains to illustrate the generalizability of our approach. We use the Multi-Modal CelebA-HQ \cite{CelebAMask-HQ, xia2021tedigan} dataset to train our model on the domain of human faces. This dataset consists of 30,000 images along with 10 textual descriptions for each image. We follow the default train/test split, using 6000 images for testing and the remaining for training. For the birds domain, we use the CUB Birds dataset \cite{WahCUB_200_2011},  which contains 11,788 images in total, including 2933 images for testing, along with 10 captions for each image. Finally, for the domain of cat faces, we use the AFHQ-Cats dataset \cite{choi2020starganv2} which contains a total of 5653 images, including 500 for testing. The captions for this dataset are generated using the approach mentioned in \cite{nie2021controllable} leveraging the CLIP \cite{radford2021learning} model.

\subsection{Training Details}
We use two pre-trained models trained on our datasets: StyleGAN2 generator and $e4e$ encoder. Keeping the weights of these models frozen, we train CLIPInverter using the cyclic adversarial training scheme described in the previous section. The mismatching captions are sampled in such a way that matching caption for an image is sampled 25\% of the time during training. In our experiments, for the CLIPAdapter, we empirically set $\lambda_1 = 1.0$, $\lambda_2 = 0.6$, $\lambda_3 = 0.1$, $\lambda_4 = 0.005$, $\lambda_5 = 1.0$, $\lambda_6 = 1.0$ and the learning rate to $0.0005$. For CLIPRemapper, we increase the weight of the identity loss to $\lambda_3~=~0.5$, and totally exclude the regularization loss during training.
We initialize the linear coefficient $\alpha_i$'s with 0.05 and train them together with the parameters of CLIPRemapper. %
We train CLIPAdapter for 200k iterations on a single Tesla v100 GPU, which takes about 6~days and CLIPRemapper for 20k iterations which takes about a day.%

\begin{figure*}[!t]
  \centering
  \includegraphics[width=\linewidth]{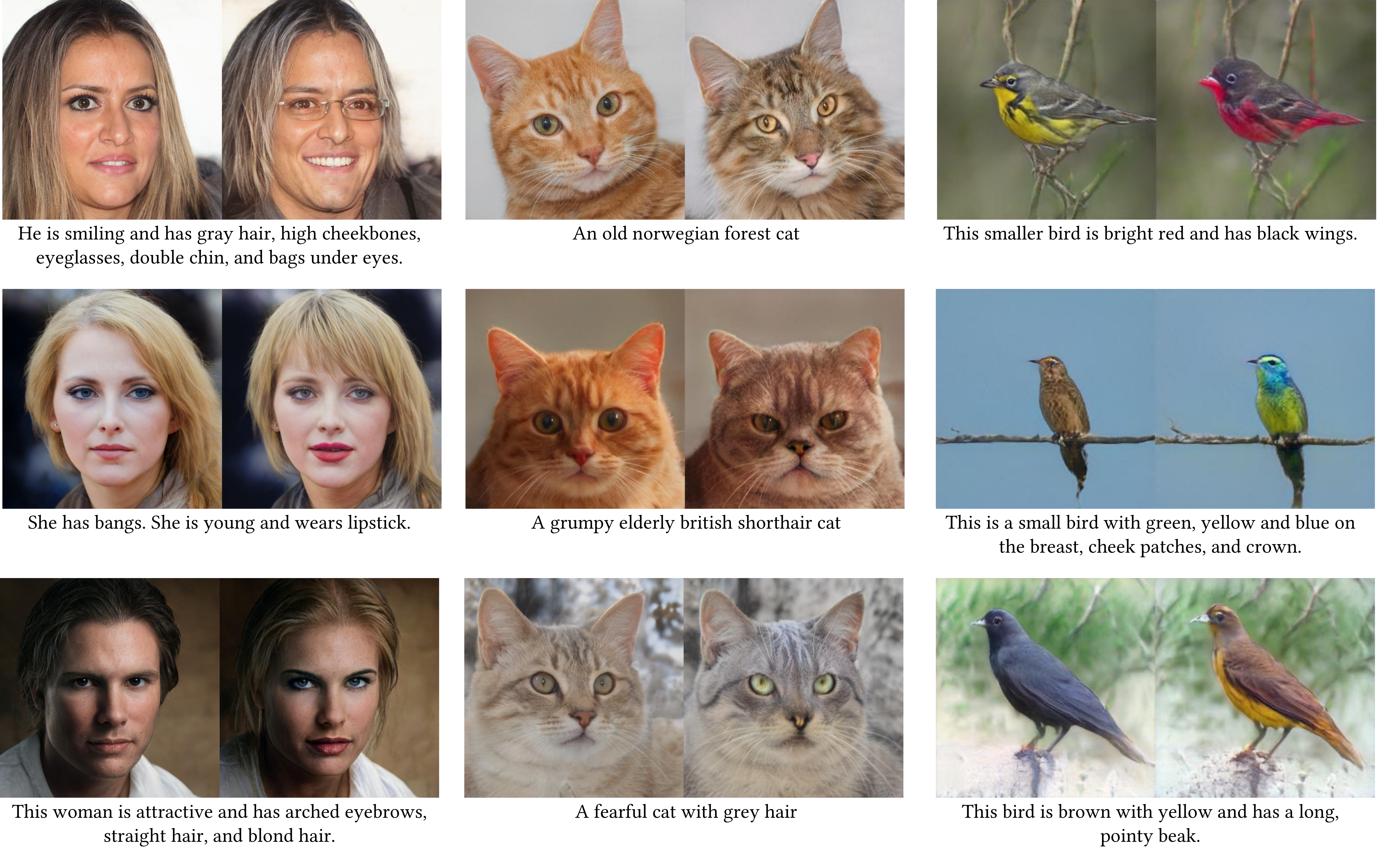}
  \vspace{-0.5cm}
  \caption{\textbf{Qualitative manipulation results.} We show sample text-guided manipulation results on human faces (\textit{left}), cat images (\textit{middle}), and bird images (\textit{right}). Our approach successfully makes local semantic edits based on the target descriptions while keeping the generated outputs faithful to the input images. The images displayed on the left side are the inversion results obtained with the e4e encoder.}
  \label{fig:qualitative}
\end{figure*}

\subsection{Evaluation Metrics}
\label{section:metrics}
Quantitative analysis of the language-guided image manipulation task is a challenging matter. The quality and the photorealism of the generated images can be evaluated with Fréchet Inception Distance (FID) ~\cite{FID}. However, there is no established way to evaluate the manipulation accuracy of a model. It is crucial that an effective model should only alter the attributes specified in the target text prompt, while preserving the original attributes for the rest of the input image. Hence, we also use the ID similarity~\cite{DBLP:conf/cvpr/DengGXZ19} to assess the identity preservation.

To evaluate the model accuracy in terms of these aspects, we propose two metrics: Attribute Manipulation Accuracy (AMA) and CLIP Manipulative Precision (CMP). Attribute Manipulation Accuracy measures how accurately a model can apply single attribute manipulations. For face images, we train an attribute classifier using the images and their attribute annotations from the CelebA~\cite{liu2015faceattributes} dataset, following~\cite{nie2021controllable}. In terms of the validation accuracy of the classifier on different attributes, we select 15 of the best performing attributes, such as \textit{blond hair}, \textit{chubby}, \textit{mustache} (see the appendix for the full list of attributes), out of 40 that are included in CelebA. Here, we have two versions of the AMA score. AMA-Single measures the accuracy of single attribute manipulations using our model. To evaluate this, we generate 50 image manipulations for each of the 15 selected attributes, resulting in a total of 750 images. For each manipulation, we employ pre-defined text prompts that specifically mention the attribute of interest, such as \textit{``This person has blond hair''}. The accuracy is then determined by assessing how well the generated images align with the intended attribute manipulation. We evaluate the accuracy of these manipulations using the attribute classifier and take the mean of the accuracy across all the attributes to obtain the final AMA score for that model. AMA-Multiple evaluates the accuracy of multiple attribute manipulations achieved by our model. We generate target descriptions that involve combinations of two or three attributes and perform 50 image manipulations for each combination, resulting in a total of 350 images. We consider the manipulation successful only when the resulting changes can be accurately classified by the corresponding attribute classifiers. In this context, a classification is deemed correct if the attribute score surpasses a threshold of 0.90.

\begin{figure}[!t]
  \centering
  \includegraphics[width=\linewidth]{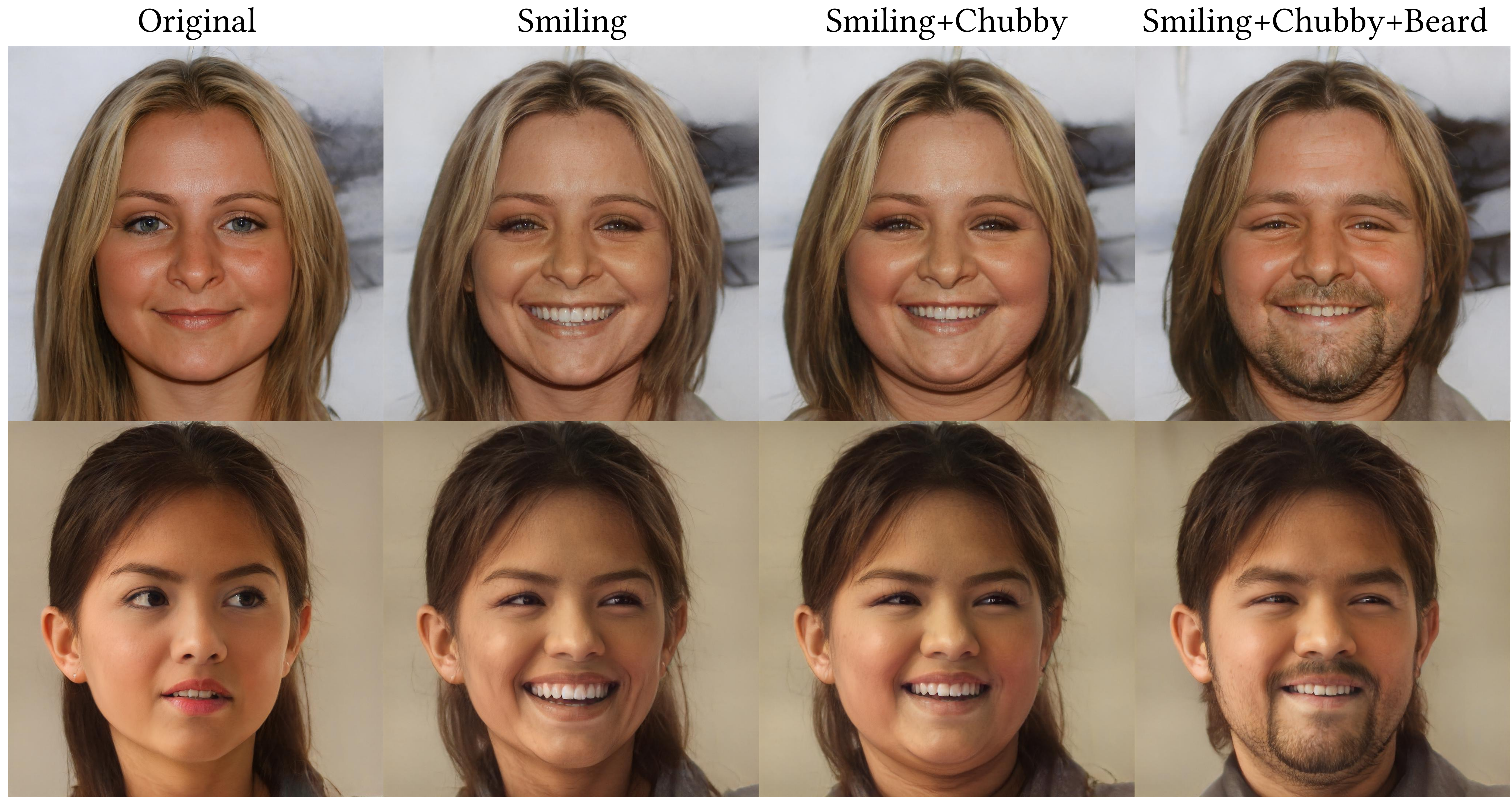}
  \caption{\textbf{More qualitative results.} We provide example manipulation results where we apply various compositions of several facial attributes as target descriptions.}
  \label{fig:composition}
\end{figure}

For cat and bird images, we use CLIP as a zero-shot classifier to calculate the AMA. We employ 30 attributes present in the AFHQ-Cats~\cite{choi2020starganv2} and sample 40 attributes out of the 273 attributes present in the CUB~\cite{WahCUB_200_2011} dataset. For each selected attribute, we generate template based captions covering all the classes in the category that the attribute belongs to. Then, we prompt CLIP with the output image and the generated captions to obtain similarity scores for each caption. The manipulation then is successful if the caption with the correct label has the highest probability after the softmax operation on the similarity scores.

CLIP Manipulative Precision is a modified version of the Manipulative Precision metric proposed by ManiGAN~\cite{li2020manigan} that uses the pre-trained CLIP~\cite{radford2021learning} image and text encoders. CMP measures how aligned the synthesized image is with the target text prompt $\mathbf{t_{target}}$ and how well the original contents of the input image are preserved. It is defined as
\begin{equation}
    \text{CMP} = (1 - \text{diff}) * \text{sim},
    \label{eq:cmp}
\end{equation}
where diff is the $\mathcal{L}_{\mathrm{1}}$ pixel difference between the input image $\mathbf{x_{in}}$ and the output image $\mathbf{x_{out}}$, and sim is the CLIP similarity between the output image $\mathbf{x_{out}}$ and the target textual description $\mathbf{t_{target}}$. We calculate the CMP for each of the images generated for the AMA score and take their average to obtain the final CMP score for the corresponding model.

\subsection{Qualitative Results}

\begin{figure*}[!t]
  \centering
  \includegraphics[width=\linewidth]{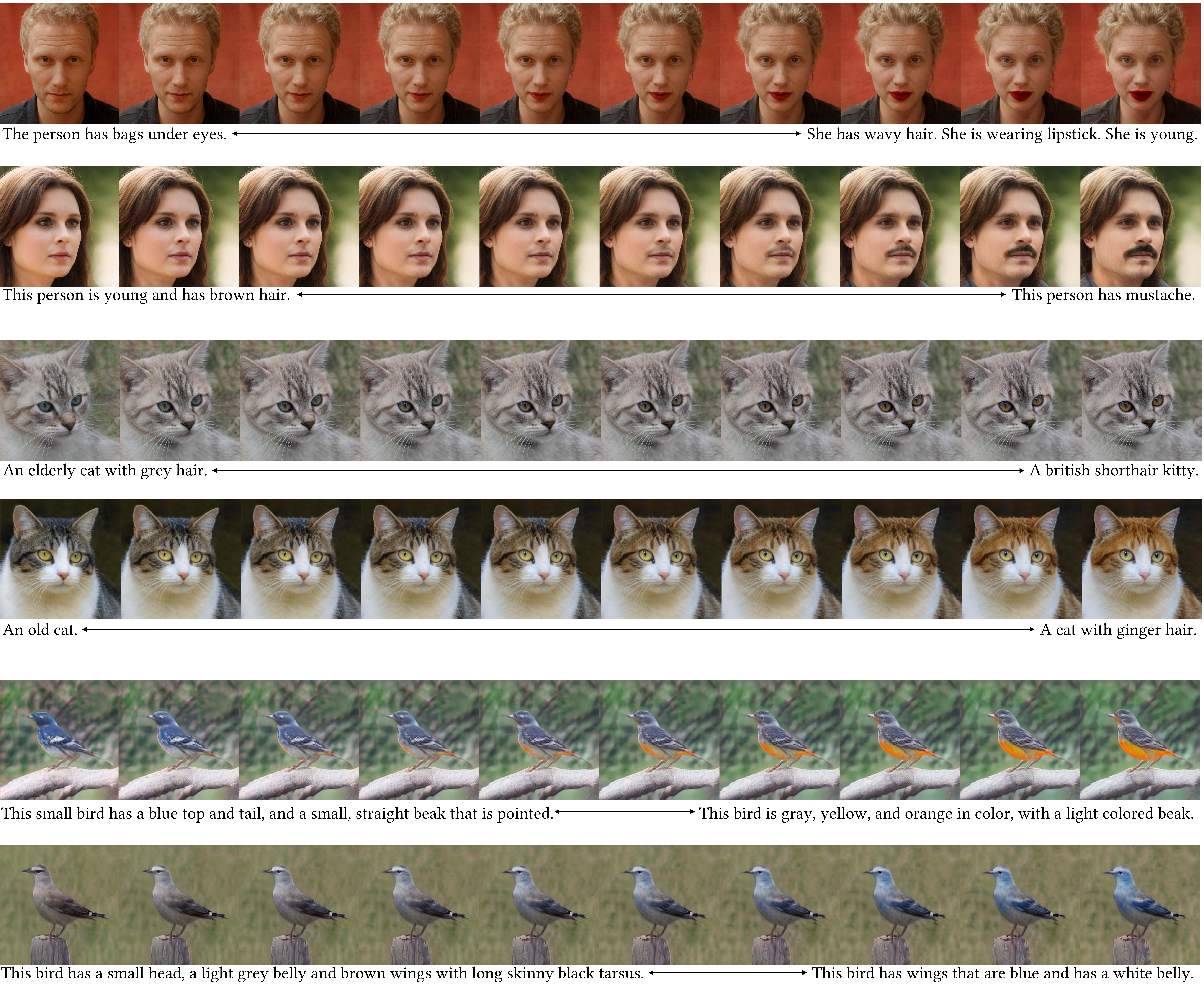}
  \caption{\textbf{Continuous manipulation results.} We show that starting from the latent code of the original image and walking along the predicted residual latent codes, we can naturally obtain smooth image manipulations, providing control over the end result. For reference, we provide the original (\textit{left}) and the target descriptions (\textit{right}) below each row.}
  \label{fig:interpolations}
\end{figure*}

In Fig.~\ref{fig:qualitative}, we show that our method can manipulate images from very different domains such as human faces, cats, and birds. Given an input image, we manipulate it by just providing a natural textual description highlighting the desired edits. As can be seen in the figure, the target descriptions can specify more than one attribute. For instance, one can simultaneously apply lipstick while changing the hair style of a woman, or can alter the attitude and appearance of a cat at the same time. 

Our method can give plausible results independent of the complexity of the provided target description. For instance, in Fig.~\ref{fig:composition}, we present the outcomes of our approach obtained by taking into account compositions of different visual attributes. They demonstrate that our method can deal with the provided compositions, and make the necessary changes in the original input images mentioned in the descriptions to its full extent.

\begin{figure*}[!t]
  \centering
  \includegraphics[width=\linewidth]{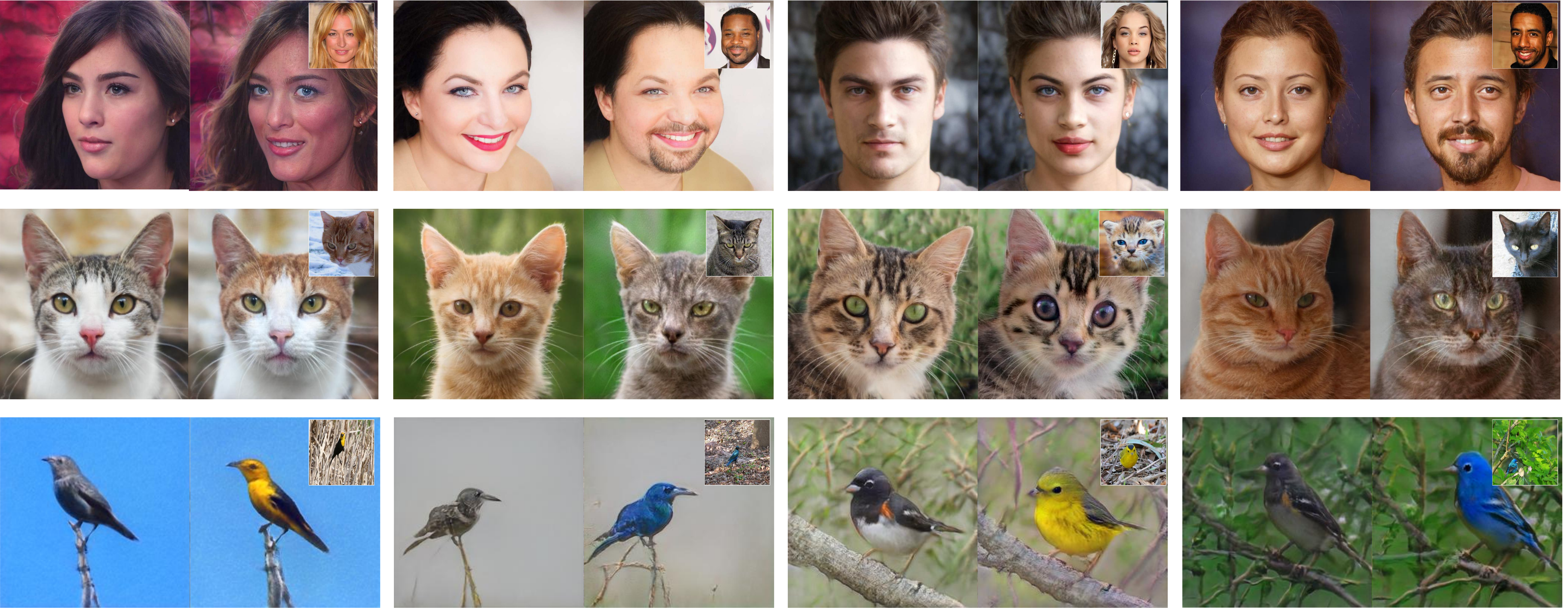}
  \caption{\textbf{Image-based manipulation results.} Our framework allows for using a reference image as the conditioning input for editing. In the figure, these reference images are given at the top-right. Results on different domains illustrate that our model can transfer the look of the conditioning images to the provided input images.}
  \label{fig:image-conditioned}
\end{figure*}

In Fig.~\ref{fig:interpolations}, we demonstrate that predicting residual latent code for a given target description has the advantage that one can continuously interpolate between the original image and the final result, which allows users to have control over the degree of changes made during the manipulation process. For example, the appearance of the subjects smoothly changes to reflect the increase in the intensity of the lipstick, and the color of the cats and the bird slightly changes.

\begin{figure}[!t]
  \centering
  \includegraphics[width=\linewidth]{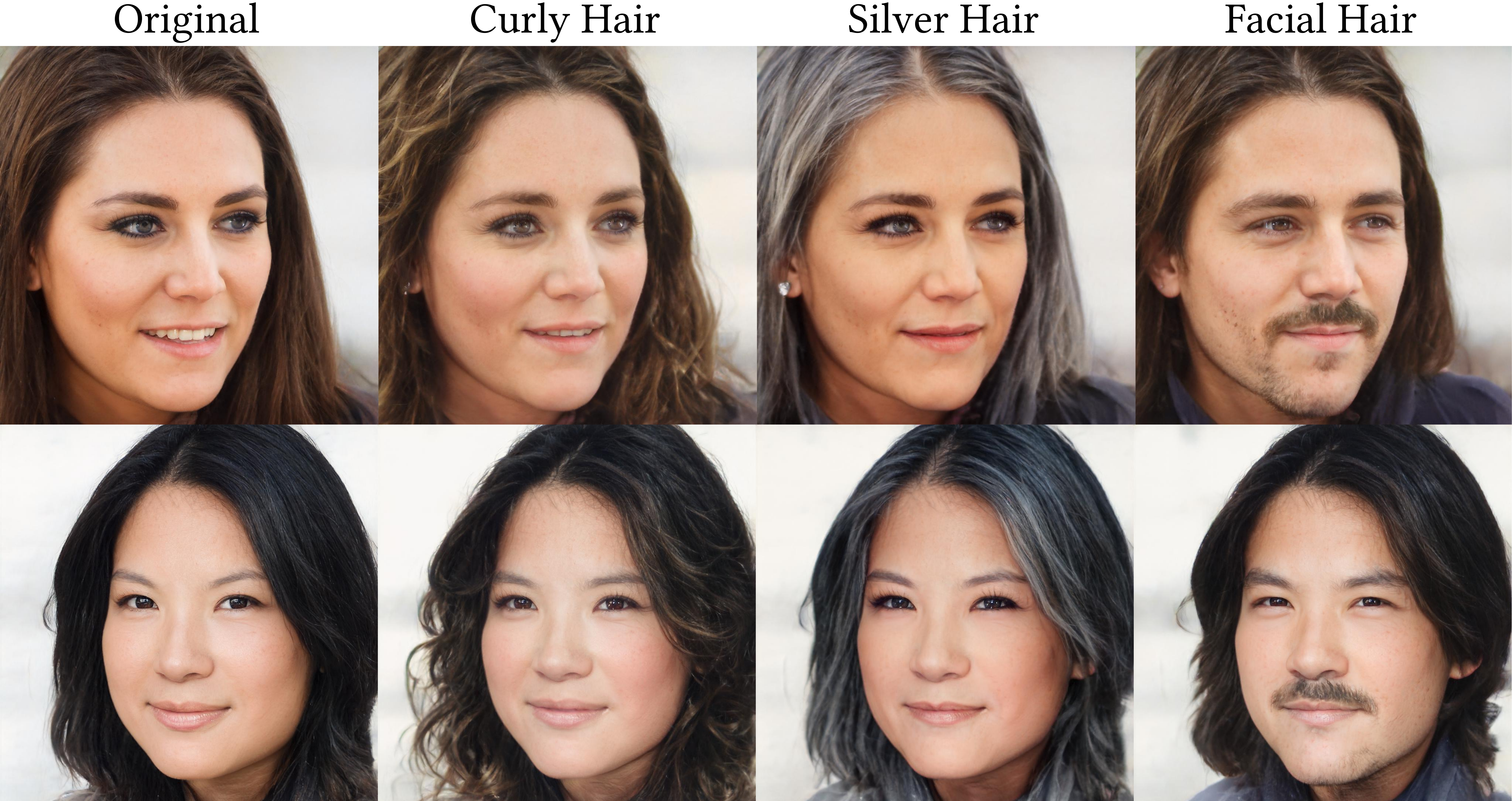}
  \caption{\textbf{Additional manipulation results with out-of-distribution training data.} We demonstrate that our CLIPInverter method can perform manipulations with target descriptions involving words never seen during training but semantically similar to the observed ones.}
  \label{fig:zeroshot}
\end{figure}

To some extent, our approach can also perform edits in a zero-shot setting by using descriptions never seen during training. The key to this ability lies in the use of the CLIP-based text guided adapters which enable to align the visual and the textual domains and map out of domain textual descriptions to a semantic editing direction in the latent space. Hence, even the terms in the target descriptions have not been observed for the first time, our method can make the necessary changes in the input images if semantically similar terms have been seen during training. For instance, in Fig.~\ref{fig:zeroshot}, we include a number of cases where the color or the structure of the hair is manipulated using novel 
descriptions that do not exist in the training set such as \emph{curly hair}, \emph{silver hair}, and \emph{facial hair}. 

In our proposed CLIPAdapter, we employ CLIP embeddings of the text prompt to modulate the convolutional feature maps to predict the residual latent code, representing the changes on the input image required to meet the desired target description. In fact, CLIP model learns the alignment between images and text via a contrastive learning objective and discovers a common semantic space. Hence, our framework also allows for using exemplar images as the conditioning element without any changes or training. In Fig.~\ref{fig:image-conditioned}, we provide some qualitative results for such image-based manipulations performed by our proposed approach. We observe that although no further training is done by considering reference images instead of target description, our model achieves a good performance on transferring the appearance of the provided reference images to the input images. 

We refer readers to the supplementary material for more manipulation results.

\begin{figure*}[!t]
  \centering
  \includegraphics[width=0.925\linewidth]{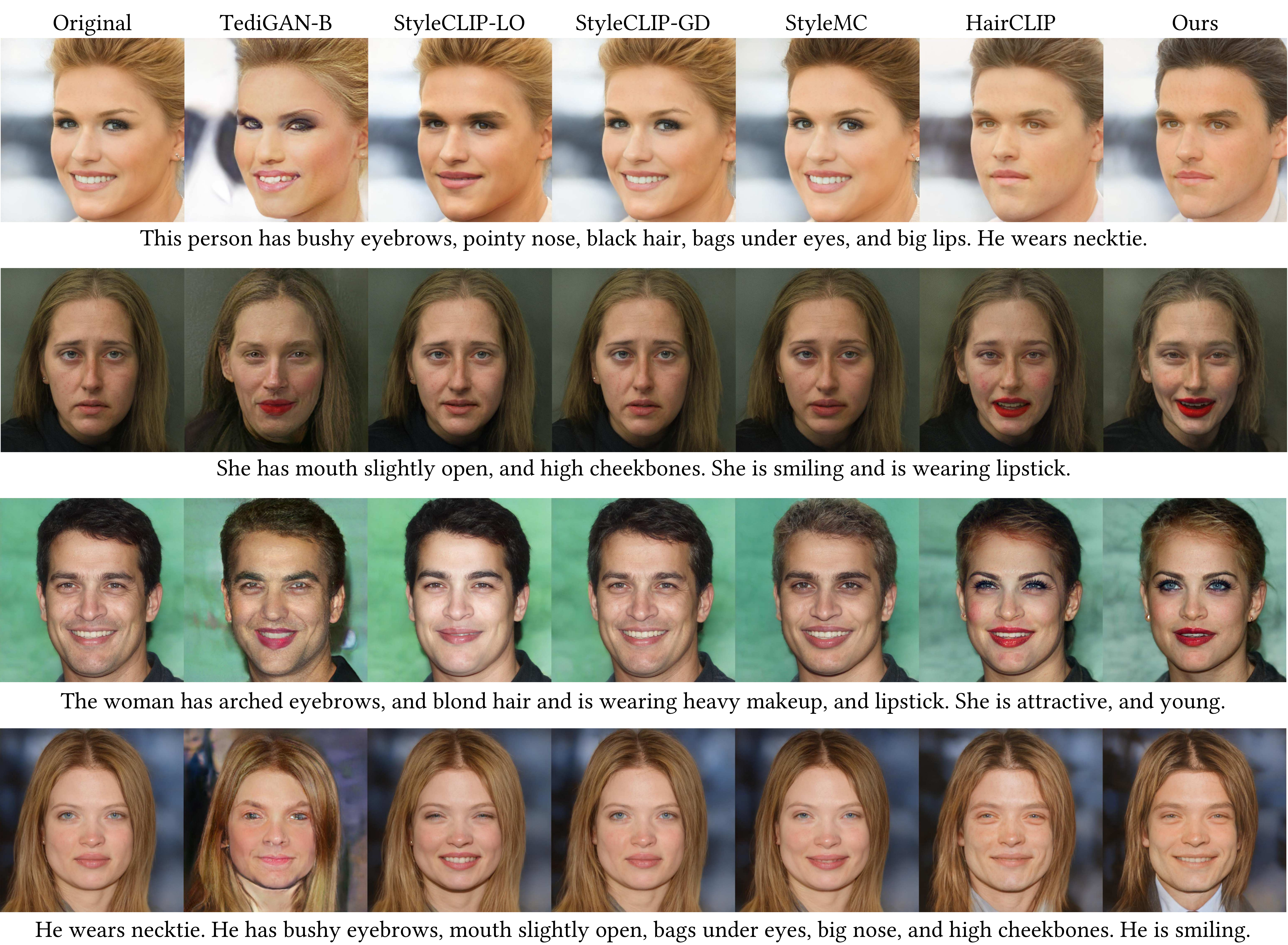}
  \caption{\textbf{Comparison against the state-of-the-art text-guided manipulation methods.} Our method applies the target edits mentioned in the given descriptions much more accurately than the competing approaches, especially when there are multiple attributes present in the descriptions.}
  \label{fig:comparisons}
\end{figure*}

\begin{figure*}[!t]
  \centering
  \includegraphics[width=\linewidth]{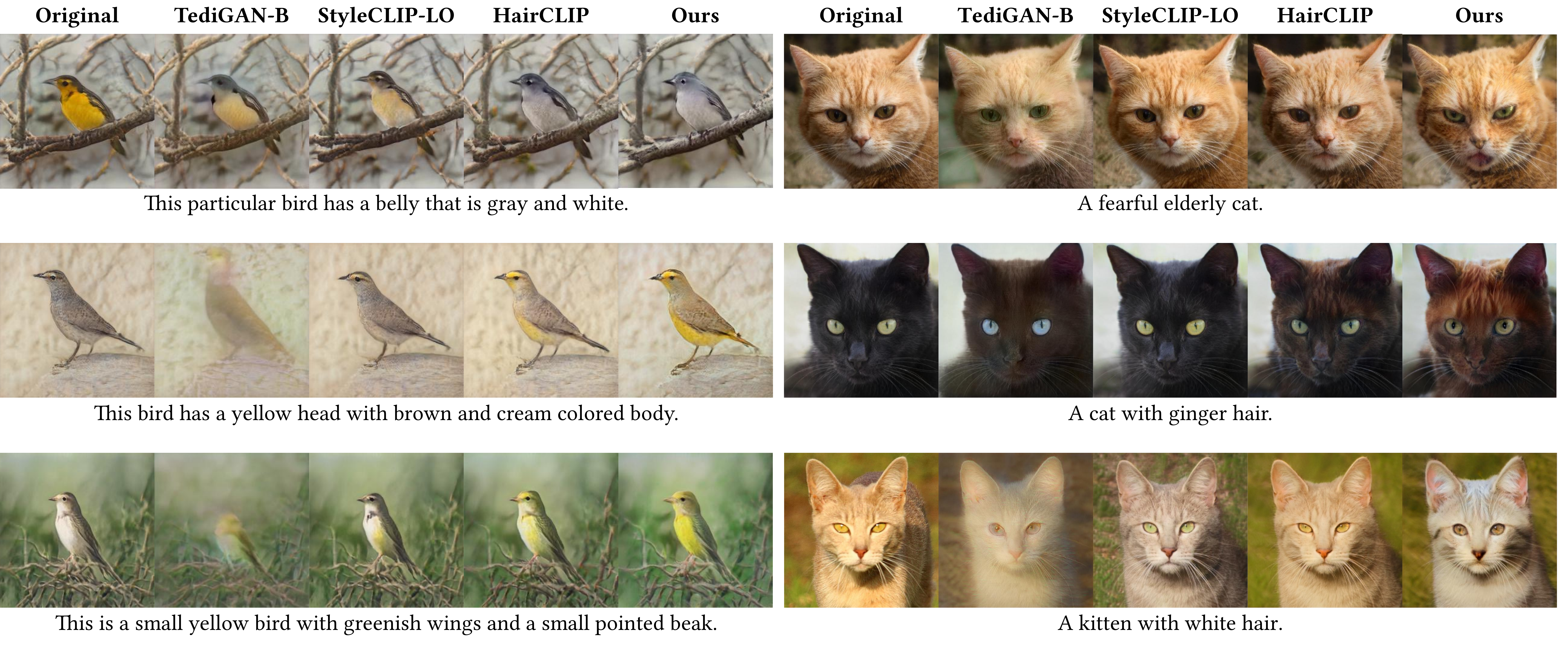}
  \caption{\textbf{Comparisons against other approaches on bird and cat images.} As compared to TediGAN, our model generates reasonable manipulation results which are more consistent with the given target descriptions.%
  }
  \label{fig:comparisonbirdcat}
\end{figure*}

\subsection{Qualitative Comparisons to Other Text-guided Manipulation Methods}

We compare our approach with various existing methods, including TediGAN \cite{xia2021tedigan}, StyleCLIP \cite{Patashnik_2021_ICCV}, StyleMC \cite{kocasari2021} and HairCLIP~\cite{wei2022hairclip}. For StyleCLIP, we use the latent optimization based model \mbox{StyleCLIP-LO}, and for TediGAN, we use the CLIP-based optimization approach (TediGAN-B). %
In all of our experiments, we use the public implementations provided by the authors. For HairCLIP, we slightly modify its neural architecture and train it accordingly. In the original paper, they do consider different conditioning vectors for the mapper modules encoding hairstyle and hair color as they refer to details from different scales. Since, we focus on a generic text-guided manipulation process where it is hard to separate the textual terms into fine, mid and high-level attributes, we let the embedding of the whole target description suggested by CLIP text encoder to condition the mappers equally. 
All of these approaches use StyleGAN2 as a frozen generator and utilize the CLIP embedding to measure the image and text similarity.

In Fig.~\ref{fig:comparisons}, we provide some qualitative comparisons between our method and the baselines on a number of human face images. As can be seen from the figure, %
our approach gives more accurate edits as compared to the existing methods, especially for captions that describe multiple attribute manipulations. For instance, for the first image, our model is able to make meaningful changes to the original input image to reflect the look depicted in the target description, and apply the gender change as well as changes in the eyebrows, hair, eyes, lips and the outfit. For the second input image, our model is able to generate the smile and the lipstick while most of the other methods fail to apply both changes at the same time. In the last two examples, our manipulation results again reflect the given target descriptions -- much better than those of the competing approaches. Our method manipulates the gender, hair color, eyebrows, age of the man and applies makeup. Similarly, it generates a smile for the woman and makes her wear a jacket, which is inline with the necktie mentioned in the description. Similarly, in Fig.~\ref{fig:comparisonbirdcat}, we compare our results with those of the TediGAN-B, StyleCLIP-LO and HairCLIP methods on bird and cat images. Like the human faces, our model is able to generate visually more pleasing and relevant results than the competing approaches. For instance, our model is able to capture the yellow-greenish color mentioned in the description for the bird in the third row and the fearful look for the cat in the first row while other methods result in poor manipulations. For birds and cats, we could not provide any comparison against StyleCLIP-GD and StyleMC as their codebase use a different implementation of the StyleGAN and they do not provide pre-trained models for these datasets. In the supplementary material, we provide additional visual comparisons.

\subsection{Quantitative Comparisons to Other Text-guided Manipulation Methods}

We quantitatively compare our approach to the same approaches that are compared in the qualitative comparisons, namely  TediGAN~\cite{xia2021tedigan}, StyleCLIP-LO and StyleCLIP-GD~\cite{Patashnik_2021_ICCV}, StyleMC~\cite{kocasari2021} and HairCLIP~\cite{wei2022hairclip}. We use the four metrics mentioned in Section~\ref{section:metrics} (Fréchet Inception Distance (FID), Attribute Manipulation Accuracy (AMA), CLIP Manipulative Precision (CMP)) and Identity Similarity (ID) for these quantitative comparisons. The official PyTorch implementation~\cite{Seitzer2020FID} is utilized to calculate the FID scores. The AMA and the CMP scores are calculated using the procedure described in Section~\ref{section:metrics}.

Table~\ref{tab:quantitative_celeba} shows the quantitative comparisons for our model against various state-of-the-art approaches. TediGAN-B achieves fairly good FID and CMP scores. However, from the qualitative results, we observed that TediGAN-B exploits adversarial ways to optimize the CLIP similarity without changing the input pixels much while failing to apply the manipulations and producing distorted images. 

While performing well in terms of either one or two metrics, the competing approaches usually fail to be competitive across all four metrics. StyleCLIP-LO is able to achieve a fairly comparable CMP, since it optimizes the CLIP similarity for each instance, and a good FID score but fails to apply the given attribute manipulations accurately. StyleMC also achieves a good FID score since it finds directions in the $\mathcal{S}$ space. However, it also fails to output accurate manipulations. Even though StyleCLIP-GD performs better than these two models, its performance still falls behind the performance of our approach. Finally, HairCLIP achieves the best scores out of the competing approaches. The results demonstrate the superiority of our model against HairCLIP, as our method achieve much higher manipulation accuracies while remaining competitive in terms of the FID and ID scores. Our approach finds a good balance for the distortion and editability problem by applying manipulations successfully while being comparable in terms of photorealism. Hence, they are able to achieve good scores across all four metrics. %

\begin{table}[!t]
    \caption{\textbf{Quantitative comparisons on the CelebA dataset.} Our approach exhibits superior manipulation accuracy compared to other methods, particularly for manipulations involving multiple attributes, while maintaining a comparable level of perceptual quality. The best and second-best performing models are highlighted in bold and underlined, respectively.}
    \resizebox{\linewidth}{!}{  
\begin{tabular}{l|ccccc}
        \toprule
             & \multicolumn{1}{c}{FID~$\downarrow$} & \multicolumn{1}{c}{CMP~$\uparrow$} & \multicolumn{1}{c}{AMA (Single)~$\uparrow$} & \multicolumn{1}{c}{AMA (Multiple)~$\uparrow$} & \multicolumn{1}{c}{ID~$\uparrow$} \\
             
        \midrule
TediGAN-B & \textbf{55.424} & \textbf{0.285} & 11.286 & 1.142 & 37.97 \\
StyleCLIP-LO & \underline{80.833}                   & 0.210                  & 15.857                         & 3.429 & 29.69                              \\
StyleCLIP-GD & 82.393                   & 0.191                  & 33.143                            & 11.429 & \underline{57.37}                            \\
StyleMC      & 84.088                   & 0.187                  & 12.143                           & 2.857 & 30.05                             \\
HairCLIP     & 93.523                   & 0.218                  & \underline{41.571}                            & \underline{15.143} & \textbf{57.50}                              \\
Ours         & 97.210                        & \underline{0.221}                       & \textbf{61.429}                                  & \textbf{41.714} & 52.14  \\
         \bottomrule
\end{tabular}
}
\label{tab:quantitative_celeba}
\end{table}

\begin{table}[!t]
    \caption{\textbf{Quantitative comparisons on the AFHQ-Cats and CUB datasets.} Our approach demonstrates superior manipulation accuracy compared to other methods, while also preserving a comparable perceptual quality. The best and second-best performing models are highlighted in bold and underlined, respectively.}
    \resizebox{\linewidth}{!}{  
\begin{tabular}{l|ccc|ccc}
        \toprule
\multicolumn{1}{c}{} & \multicolumn{3}{c}{AFHQ-Cats}                       & \multicolumn{3}{c}{CUB}    \\
        \midrule                           
                     & FID $\downarrow$            & CMP $\uparrow$             & AMA $\uparrow$             & FID $\downarrow$            & CMP $\uparrow$             & AMA $\uparrow$             \\ 
        \midrule
TediGAN-B            & 39.414          & \textbf{0.255} & \textbf{82.467} & 42.007          & \textbf{0.233} & \underline{59.500}          \\
StyleCLIP-LO         & \textbf{18.771} & 0.226          & 48.133          & \textbf{19.209} & 0.211          & 27.000          \\
HairCLIP             & \underline{21.087}          & 0.227          & 44.667          & 26.447          & 0.218          & 57.050          \\
Ours                 & 24.172          & \underline{0.245}          & \underline{76.467}          & \underline{25.837}          & \underline{0.221}          & \textbf{66.000}  \\
         \bottomrule
\end{tabular}}
\label{tab:quantitative_birdcat}
\end{table}

Table~\ref{tab:quantitative_birdcat} presents the quantitative comparisons on the AFHQ-Cats and the CUB datasets.
Since CLIP is used as a similarity metric in CMP and as a zero-shot classifier in AMA estimations, TediGAN-B %
again achieves really good scores in these two metrics.
However, as seen from the FID scores and the results shown in Fig.~\ref{fig:comparisonbirdcat}, it gives highly blurred and non-realistic outputs that are not actually in line with the target descriptions. Another optimization based method, StyleCLIP-LO, achieves worse AMA and CMP scores than TediGAN-B, but better FID. Their loss functions allow the model to output realistic outputs, but they fail to apply the manipulations successfully, which can be seen in Fig.~\ref{fig:comparisonbirdcat}. HairCLIP generates images that are better in line with the descriptions than the aforementioned methods. However, our approach outperforms HairCLIP by a large margin in terms of CMP and AMA while having a fairly close or even better FID values. We underlined the second best performing models for each metric, to demonstrate the superiority of our approach against the others, since the best performing models usually exploit adversarial ways to optimize the CLIP similarity which yield high CMP and AMA values or fail to apply the manipulations which yield better FID values.

For quantitative analysis, we conduct a user study via Qualtrics to evaluate the performance of our approach and all the other competing methods. Specifically, in this user study, we focus on two important aspects: (1) the accuracy of the edits with respect to the given target descriptions, and (2) the photorealism of the manipulated images. In our human evaluation, we randomly generate 48 questions, and divide them into 3 groups, with 16 questions each. We make sure that at least 14 different subjects answer each of these group of questions. To measure the accuracy, we show the users an input image, a target description, and the manipulation results of all of the competing methods, and ask them to rank the results against each other with respect to how consistent the edits are to the provided description. The participants perform this by dragging the images into their preferred order, where the left-most position  refers to the worst result having rank order 1 and the right-most one represents the best outcome at rank order 6. In order to avoid any bias in the evaluation, the outputs of the methods are displayed in random order at each time. For the questions regarding photorealism, we design a similar ranking task, but this time, we show all the results in random order, and ask the participants to order these results with respect to how realistic they look. Please refer to the supplementary for a screenshot of our user study given to the participants.

\begin{figure*}[!t]
 \centering
 \includegraphics[width=\linewidth]{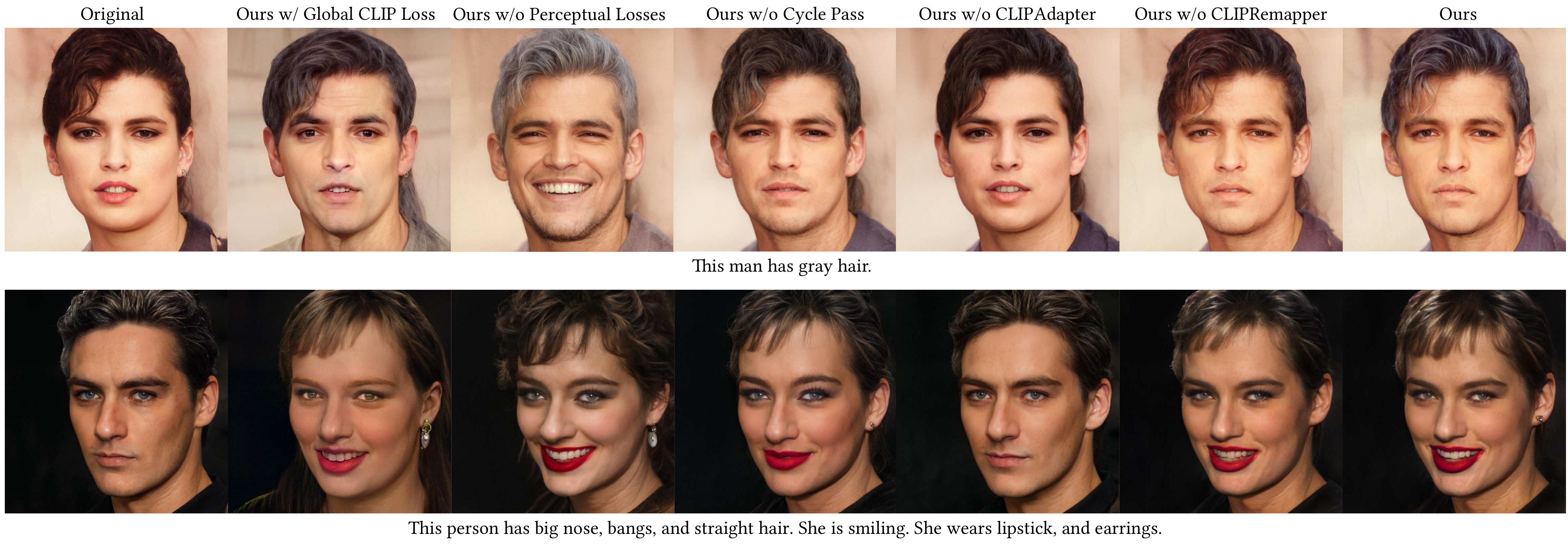}
 \caption{\textbf{Qualitative results for the ablation study}. The global CLIP loss leads to unintuitive and unnatural results. Without perceptual losses, unwanted manipulations occur. Without the cycle pass or CLIPRemapper, we are not able to apply all the desired manipulations.%
 }
 \label{fig:ablation}
\end{figure*}

\begin{table}[!t]
    \centering
    \caption{\textbf{User study results.} The table represent the average rankings of the methods with respect to accuracy and realism, where the higher the value is the better the method is. The participants favor the results of our proposed model over the current state-of-the-art when the accuracy of the manipulations is considered.}
    \resizebox{\linewidth}{!}{
    \begin{tabular}{lc@{$\;\;\,$}c@{$\;\;\,$}c@{$\;\;\,$}c@{\;\;\,}c@{\;\;\,}c}
        \toprule
        Task & TediGAN-B & StyleCLIP-LO & StyleMC & StyleCLIP-GD & HairCLIP & Ours\\
        \midrule
        Acc. & 1.848 & 3.401 & 3.526 & 3.611 & 4.015 & 4.598 \\
        Real. & 1.218 & 4.604 & 4.282 & 3.609 & 3.544 & 3.743 \\
        \bottomrule
     \end{tabular}}

    \label{tab:user_study}
\end{table}

Table~\ref{tab:user_study} summarizes the results of our study where the average ranking scores are reported. We find that in terms of the accuracy, the human subjects prefer our proposed method against all the competing approaches. That is, our method makes only the necessary edits in the input images with respect to the given target descriptions in a precise manner. HairCLIP and StyleCLIP-GD give the next most accurate results following our model. In terms of photorealism, our results are also superior than these two approaches, indicating that our results are both accurate and photo-realistic. That said, the human subjects find the photorealism of the results of the concurrent StyleMC and StyleCLIP-LO models significantly better. However, the accuracy questions indicate that both StyleMC and StyleCLIP-LO have difficulty in manipulating the given input images in regard to the target descriptions, in contrast to our proposed model. StyleMC and StyleCLIP-LO, in general, make minimal, mostly insufficient changes in the input images (as also can be seen from Fig.~\ref{fig:comparisons}), and thus do not degrade the photorealism much.

\subsection{Ablation Study}

During training our model, we leverage different loss terms. In order to analyze the contributions of these loss terms, we have performed an ablation study where we either remove or modify some of these loss terms during training. We provide visual comparisons between these models separately trained on different loss terms in Fig~\ref{fig:ablation}.

Firstly, we employ the directional CLIP loss following \cite{stylegan-nada}, to better enforce the image and description similarity. Compared to the global CLIP loss, which directly minimizes the distance between the manipulated image $\mathbf{x}_{out}$ and the text prompt $\mathbf{t}_{target}$ in the CLIP space, the directional CLIP loss aligns the directions between the real and target descriptions and input and output images. As can be seen in the second column of Fig~\ref{fig:ablation}, the global CLIP loss suffers from artificial-looking manipulations and results in poorly constructed facial attributes as compared to the directional CLIP loss. 

Secondly, in order to preserve the features and the details of the input image in the areas that we do not wish to modify, we employ the perceptual $\mathcal{L}_{\mathrm{2}}$ and the $\mathcal{L}_{\mathrm{LPIPS}}$ losses between the input and the output images. In theory, these perceptual loss terms contradict the directional CLIP loss since the CLIP loss is trying to enforce the image \& text similarity by manipulating the pixel values. In order to analyze the contribution of these perceptual terms, we have reduced the weights of these loss terms in the overall objective. The third column in Fig.~\ref{fig:ablation} shows a manipulation example from this experiment. As can be seen, the smile in the first row is also modified, and the model manipulates the hair style to curly hair in the second row even though this manipulations were not mentioned in the target description. This experiment demonstrates the necessity of these perceptual loss terms in order to prevent unwanted manipulations.

Thirdly, we employ a cyclic-adversarial training strategy, where we first manipulate the image with a mismatching caption, and then recover it by manipulating the output of the first pass with the matching target description. The fourth column in Fig.~\ref{fig:ablation} shows an example manipulation from the experiment where we remove this cyclic training regime. Even though the output is visually similar to the output from our full model, we observe that the cyclic consistency loss helps with the preservation of the identity as well as the manipulation accuracy.

Finally, we utilize a CLIP-guided correction module CLIPRemapper to apply the manipulations more accurately and increase the image fidelity. We see from the last two columns of the figure that without CLIPRemapper, the model is not able to apply all of the specified manipulations accurately, like the hair color in the first row or the earrings in the second row. %

\begin{table}[!t]
\centering
\caption{\textbf{Quantitative analysis of the ablation study.} We have performed a quantitative analysis of the ablation study where we calculated the metrics for each of the described experiments for our model. The results demonstrate that our model finds a good balance for applying manipulations without decreasing the perceptual quality of the generations. The best and second-best performing models are highlighted in bold and underlined, respectively.}
\begin{tabular}{@{}l|ccc@{}}
\toprule
                           & FID $\downarrow$                                    & CMP $\uparrow$                                      & AMA $\uparrow$                                     \\ \midrule
Ours w/ Global CLIP Loss   & \textbf{83.404} & \textbf{0.221}                                  &  25.429 \\
Ours w/o Perceptual Losses & 105.432                                  &  0.194 &  \textbf{65.571}                                \\
Ours w/o Cycle Pass        & \underline{85.851}                                  &      0.215       &  40.857                                  \\
Ours w/o CLIPAdapter                       & 89.244                                  &   0.202                               &          41.571                          \\ 
Ours w/o CLIPRemapper                       & 88.395                                  &   \underline{0.216}                               &          53.28                          \\ 
Ours                    &    97.210                                 &                            \textbf{0.221}     &   \underline{61.429}                              \\ \bottomrule
\end{tabular}
\label{tab:ablation}
\end{table}

Table~\ref{tab:ablation} shows the quantitative analysis of the experiments described above. The metrics verify that the global CLIP loss performs much worse in terms of attribute manipulations. This model is able to achieve a high CMP since it directly optimizes image and text similarity in the CLIP space, rather than aligning semantic directions. When we reduce the weight of the perceptual losses, the model is able to apply the manipulations with very high accuracy. However, this comes with the price of perceptual quality as FID suggests, and unwanted manipulations as CMP suggests. Adding the cycle pass gives us a better supervision signal to train the model, as the improvements in the accuracy and the CMP suggest. Without CLIPAdapter, our model is not able to achieve great accuracy scores, suggesting that the adapter layers yield the residual latent codes corresponding to semantic directions successfully. Finally, adding CLIPRemapper to our model highly boosts the manipulation performance, with slightly decreasing photorealism in terms of FID. Overall, these quantitative results demonstrate that the combination of the loss functions and the light-weight modules we use allows our model to perform well across all metrics and apply manipulations accurately while preserving the photorealism and preventing unwanted changes. %

\subsection{Limitations}

\begin{figure}[!t]
  \centering
  \includegraphics[width=\linewidth]{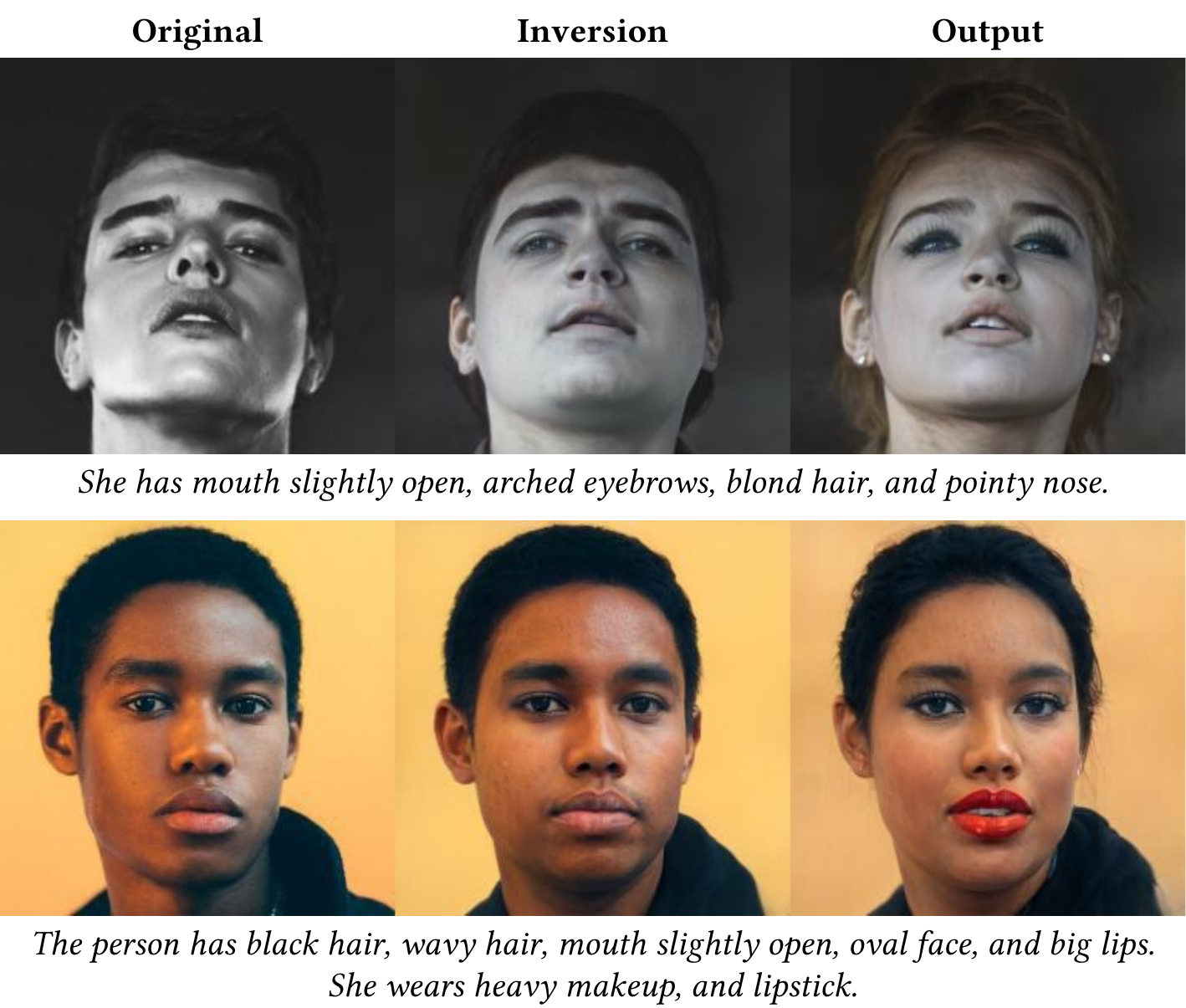}
  \caption{\textbf{Limitations due to GAN inversion step.} We present editing results for various challenging inputs (\textit{left}) where the inputs have different lighting conditions, shadows then the training data. The underlying inversion model struggles to capture these details in the initial inversion phase (\textit{middle}). Our approach is able to generate consistent results (\textit{right}) with the reconstructed images.}
  \label{fig:limitations2}
\end{figure}

In our approach, the CLIPAdapter module can be integrated with any inversion network model. Consequently, the limitations of the underlying inversion network are inherited by our approach. For instance, when employing the $\text{e4e}$, the network may struggle to find accurate latent codes for inputs with unusual poses or challenging lighting conditions. Hence, the reconstructions occasionally result in alterations to identity and the loss of certain details. Despite these limitations, our approach remains capable of generating outputs that align with the provided textual descriptions. It is important to note that the output images are consistent with the reconstructed images, rather than the input images themselves. Fig.~\ref{fig:limitations2} demonstrates manipulation results using our approach with various challenging inputs. As observed, our method successfully applies the manipulations with respect to the desired reconstructions, however, some details present in the input images, such as shadows or specific lighting conditions, may not be fully preserved during the unconditional inversion phase. It is worth mentioning that this is a common limitation of current GAN-based editing methods, as many of these approaches rely on a pre-trained encoder like $\text{e4e}$ to obtain an initial inversion of the inputs. For a comprehensive comparison of competing approaches under these challenging cases, we refer readers to the supplementary material.

\begin{figure}[!b]
  \centering
  \includegraphics[width=\linewidth]{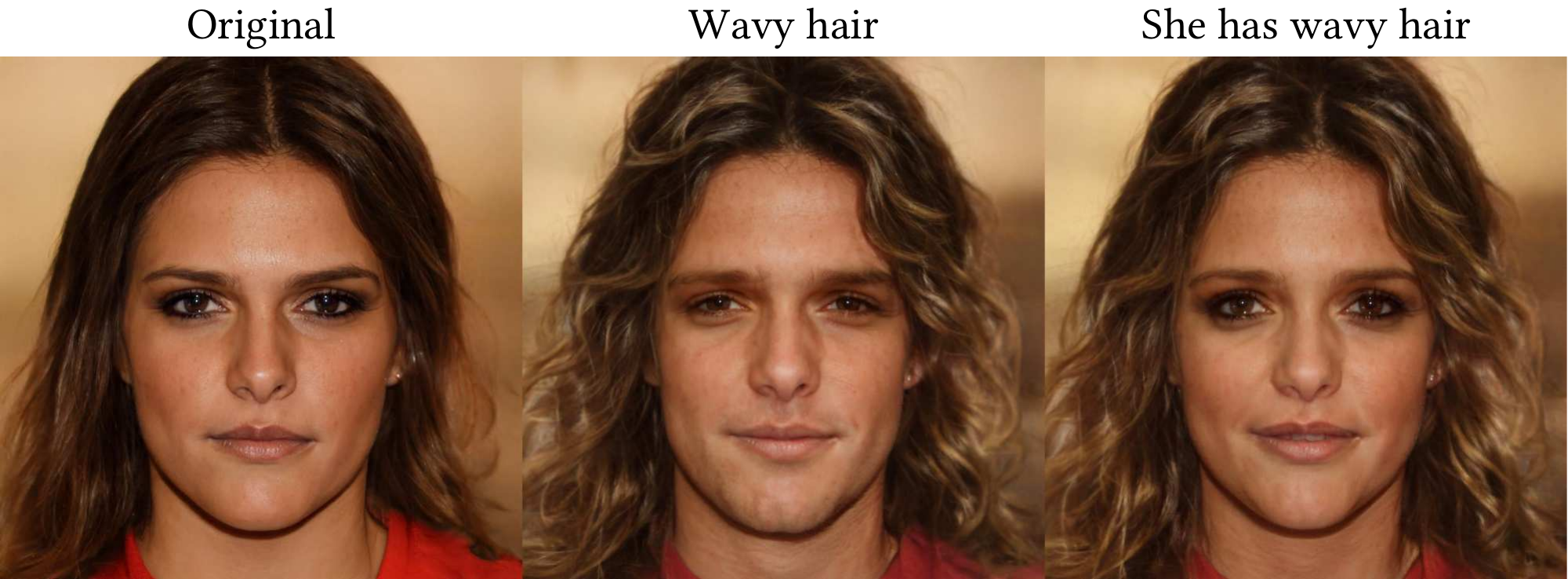}
  \caption{\textbf{Limitations of our proposed CLIPInverter method.} Our approach might make some undesired changes to the given input image not mentioned in the provided textual description due to the biases that exist in the training set. This problem can be prevented by providing more comprehensive descriptions.}
  \label{fig:limitations}
\end{figure}

Additionally, the effectiveness of our method mainly lies in the proposed text-guided image encoder CLIPInverter, which estimates the residual latent code to capture the desired changes. Since CLIPInverter is trained by using a set of training images paired with corresponding textual descriptions, we observe that results of our approach might be affected by the biases that exist in the training data. For instance, Multi-Modal CelebA-HQ dataset containing human face images consists of 10 descriptions for each image, but we observe that the descriptions are not diverse, often using similar adjectives referring to certain attributes.
Moreover, there is an imbalance between the number of female and male images, causing a bias towards a specific gender in certain attributes. When only attributes are used in the textual descriptions without any pronouns, unexpected gender manipulations might occur due to these biases. As observed in Fig. ~\ref{fig:limitations}, when we only use the description ``\emph{wavy hair}'', a gender manipulation also occurs. We can alleviate this problem by using more comprehensive textual descriptions, including additional details such as ``\emph{She has wavy hair}'', which yields a much more accurate manipulation. It is an interesting future direction to tackle the bias problem in a more systematic manner.

\section{Conclusion}
In this work, we have introduced CLIPInverter, a novel text-driven image editing approach. It can be used to manipulate an input image through the lens of StyleGAN latent space solely by providing a target textual description, which is much more intuitive than the commonly-used user inputs such as sketches, strokes or segmentation masks. The key component of our approach is the proposed text-guided adapter module called CLIPAdapter, which modulates image feature maps during the inversion to extract semantic edit directions with respect to the provided target description. Moreover, we suggest a text-guided refinement module that we refer to as CLIPRemapper, which performs an additional correction step on the predicted latent code from CLIPAdapter to further boost the accuracy of the performed edits in the input image. Our model does not require an instance-level latent code optimization or a separate training for specific text prompts as done in the prior work, and thus provides a faster alternative to the approaches that exist in the literature.

Our approach is not limited to a specific domain in that it only needs a pretrained StyleGAN model. As our experimental analysis on several different datasets illustrate, our model can handle the semantic edits through textual descriptions for very different domains. Moreover, thanks to the shared semantic space provided by the CLIP \cite{radford2021learning} model between images and text, our model can also be used to perform manipulations conditioned on another image or a novel textual description that has not been seen during training. Our experiments demonstrate significant improvements over the previous approaches in that our model can manipulate images with high accuracy and quality for any description. 

Furthermore, it is important to highlight that our proposed framework is not limited to StyleGAN and can be seamlessly integrated into other deep generative models that operate on a latent space representation. Although our current implementation focuses on StyleGAN, the key contributions of our framework, namely CLIPAdapter and CLIPRemapper, are not specific to StyleGAN and can be easily adapted to other GAN architectures. This flexibility opens up opportunities for leveraging our framework in conjunction with recent advancements in latent space extension, such as dual-space GANs which exhibit enhanced disentanglement of style and content information~\cite{kwon2021iccv,xu2022cvpr}. By incorporating our framework into these models, we can further enhance manipulation accuracy and broaden the range of images that can be generated based on textual descriptions.

\begin{acks}
This work has been partially supported by AI Fellowships to A. C. Baykal and A. Basit Anees provided by the KUIS AI Center, by BAGEP 2021 Award of the Science Academy to A. Erdem, and by an Adobe research gift. 
\end{acks}

\bibliographystyle{ACM-Reference-Format}
\bibliography{bibliography.bib}

\end{document}


\title{Supplementary Material: CLIP-Guided StyleGAN Inversion for Text-Driven Real Image Editing}

\author{Ahmet Canberk Baykal}
\email{abaykal20@ku.edu.tr}
\orcid{0000-0002-0249-5858}
\affiliation{%
  \institution{Koç University}
  \country{Turkey}
}

\author{Abdul Basit Anees}
\email{aanees20@ku.edu.tr}
\orcid{0000-0003-1293-1796}
\affiliation{%
  \institution{Koç University}
  \country{Turkey}
}

\author{Duygu Ceylan}
\email{duygu.ceylan@gmail.com}
\affiliation{%
  \institution{Adobe Research}
  \country{United Kingdom}
}

\author{Erkut Erdem}
\email{erkut@cs.hacettepe.edu.tr}
\orcid{0000-0002-6744-8614}
\affiliation{%
  \institution{Hacettepe University}
  \country{Turkey}
}

\author{Aykut Erdem}
\email{aerdem@ku.edu.tr}
\orcid{0000-0002-6280-8422}
\affiliation{%
  \institution{Koç University}
  \country{Turkey}
}

\author{Deniz Yuret}
\email{dyuret@ku.edu.tr}
\orcid{0000-0002-7039-0046}
\affiliation{%
  \institution{Koç University}
  \country{Turkey}
}

\renewcommand{\shortauthors}{Baykal et al.}

\ccsdesc[500]{Computing methodologies~Image manipulation}
\ccsdesc[500]{Computing methodologies~Neural networks}

\maketitle

The purpose of this document is to provide extra material to complement the paper. Section~\ref{sec:metrics} provides detailed information on how we estimated the AMA scores in our experiments. In Section\ref{sec:user-study}, we present a brief overview of the user study conducted to subjectively assess the outcomes of the evaluated models. Section~\ref{sec:differentencoders} investigates the impact of the encoder network used in conjunction with the proposed CLIPAdapter module on editing performance. Furthermore, Section~\ref{sec:imagecondinversion} showcases experiments where CLIPAdapter is utilized for the sole purpose of inverting real images. In Section~\ref{sec:more_results}, we present comprehensive qualitative results across a wide range of domains. Additionally, Section~\ref{sec:more_comparisons} offers additional visual comparisons between the proposed CLIPInverter model and existing text-driven image editing approaches. Section~\ref{sec:diffusion_comparison} provides qualitative and quantitative comparisons against diffusion-based text-guided manipulation methods. Furthermore, Section~\ref{sec:unseen_comparison} presents comparisons with HairCLIP, an approach similar to ours, using text prompts that were not encountered during training. Section~\ref{sec:supplementary_limitations} demonstrates additional manipulation results that highlight the limitations of our framework on challenging images. Lastly, in Section~\ref{sec:clipremapper}, we specifically analyze the impact of the refinement operation performed by the CLIPRemapper module within the cats and birds domains.

\begin{figure*}[!t]
  \centering
  \setlength{\fboxsep}{0pt}%
\setlength{\fboxrule}{0.25pt}%
 \fbox{
\includegraphics[width=\linewidth]{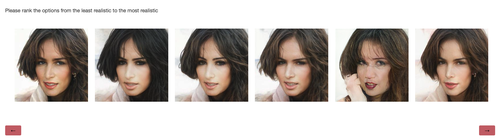}
}
  \caption{\textbf{Screenshot of a question from the user study.} The participants are asked to rank the choices in the specified order.
  }
  \label{fig:userstudyss}
\end{figure*}

\section{Details about the AMA Score}
\label{sec:metrics}
Our proposed Attribute Manipulation Accuracy (AMA) score measures how accurately a model can apply single attribute manipulations. It requires requires the availability of pretrained attribute classifiers trained for a set of attributes. As mentioned in the main paper, we hand-selected these attribute classifiers based on their robustness.

For the CelebA dataset~\cite{CelebAMask-HQ, xia2021tedigan}, we have used the following attributes in calculating the AMA score:
\begin{multicols}{2}
\begin{itemize}
    \item \textit{blonde hair}
    \item \textit{bushy eyebrows}
    \item \textit{chubby}
    \item \textit{double chin}
    \item \textit{eyeglasses}
    \item \textit{goatee}
\end{itemize}
\end{multicols}
\begin{multicols}{2}
\begin{itemize}  
    \item \textit{gray hair}
    \item \textit{heavy makeup}
    \item \textit{male}
    \item \textit{mouth slightly open}
    \item \textit{mustache}
    \item \textit{rosy cheeks}
    \item \textit{smiling}
    \item \textit{wearing lipstick}
    \item \textit{wearing necktie}
\end{itemize}
\end{multicols}

Here is the complete list of attributes we used to calculate the AMA score for the CUB dataset~\cite{WahCUB_200_2011}:
\begin{multicols}{2}
\begin{itemize}
    \item \textit{curved bill shape}
    \item \textit{blue wing}
    \item \textit{rufous wing}
    \item \textit{red wing}
    \item \textit{olive upperparts}
    \item \textit{iridescent underparts}
    \item \textit{pink underparts}
    \item \textit{blue back}
    \item \textit{rounded tail tail shape}
    \item \textit{blue upper tail}
    \item \textit{iridescent upper tail}
    \item \textit{yellow upper tail}
    \item \textit{orange upper tail}
    \item \textit{red upper tail}
    \item \textit{spotted head pattern}
    \item \textit{iridescent breast}
    \item \textit{black breast}
    \item \textit{blue throat}
    \item \textit{purple throat}
    \item \textit{pink eye}
    \item \textit{orange eye}
    \item \textit{red eye}
    \item \textit{white forehead}
    \item \textit{black under tail}
    \item \textit{rufous nape}
    \item \textit{grey nape}
    \item \textit{yellow nape}
    \item \textit{rufous belly}
    \item \textit{grey belly}
    \item \textit{black belly}
    \item \textit{broad-wings wing\\shape}
    \item \textit{long-legged-like shape}
    \item \textit{striped back pattern}
    \item \textit{spotted belly pattern}
    \item \textit{grey primary}
    \item \textit{olive leg}
    \item \textit{pink leg}
    \item \textit{white leg}
    \item \textit{purple bill}
    \item \textit{black bill}
\end{itemize}
\end{multicols}

For the AFHQ-Cats~\cite{choi2020starganv2} dataset, the set of attributes selected for the AMA score is as follows:
\begin{multicols}{2}
\begin{itemize}
    \item \textit{a kitten or kitty}
    \item \textit{an elderly or old cat}
    \item \textit{bengal cat}
    \item \textit{black}
    \item \textit{bombay cat}
    \item \textit{bored}
    \item \textit{british shorthair cat}
    \item \textit{brown}
    \item \textit{calico cat}
    \item \textit{cinnamon}
    \item \textit{cream}
    \item \textit{egyptian cat}
    \item \textit{fearful}
    \item \textit{ginger}
    \item \textit{grey}
    \item \textit{grumpy}
    \item \textit{happy}
    \item \textit{himalayan cat}
    \item \textit{maine coon cat}
    \item \textit{norwegian forest cat}
    \item \textit{persian cat}
    \item \textit{playful}
    \item \textit{ragdoll cat}
    \item \textit{savannah cat}
    \item \textit{scottish fold cat}
    \item \textit{siamese cat}
    \item \textit{siberian cat}
    \item \textit{singapura cat}
    \item \textit{snowshoe cat}
    \item \textit{white}    
\end{itemize}
\end{multicols}

\section{User Study}
\label{sec:user-study}
We have used the Qualtrics platform to perform the user study. We randomly sampled images and descriptions to generate 48 manipulations in total (24 each for accuracy and realism). These 48 questions are then divided into 3 groups of 16 questions. To present a study suitable for participants' attention span, a single participant is given only one group of questions. In each question, the participants rank the outputs of each model in the specified order. A screenshot of a question from the user study is given in Fig.~\ref{fig:userstudyss}.

\section{\textsc{CLIPAdapter} in Different Encoder Networks}
\label{sec:differentencoders}
\begin{figure}[!t]
  \centering
  \includegraphics[width=\linewidth]{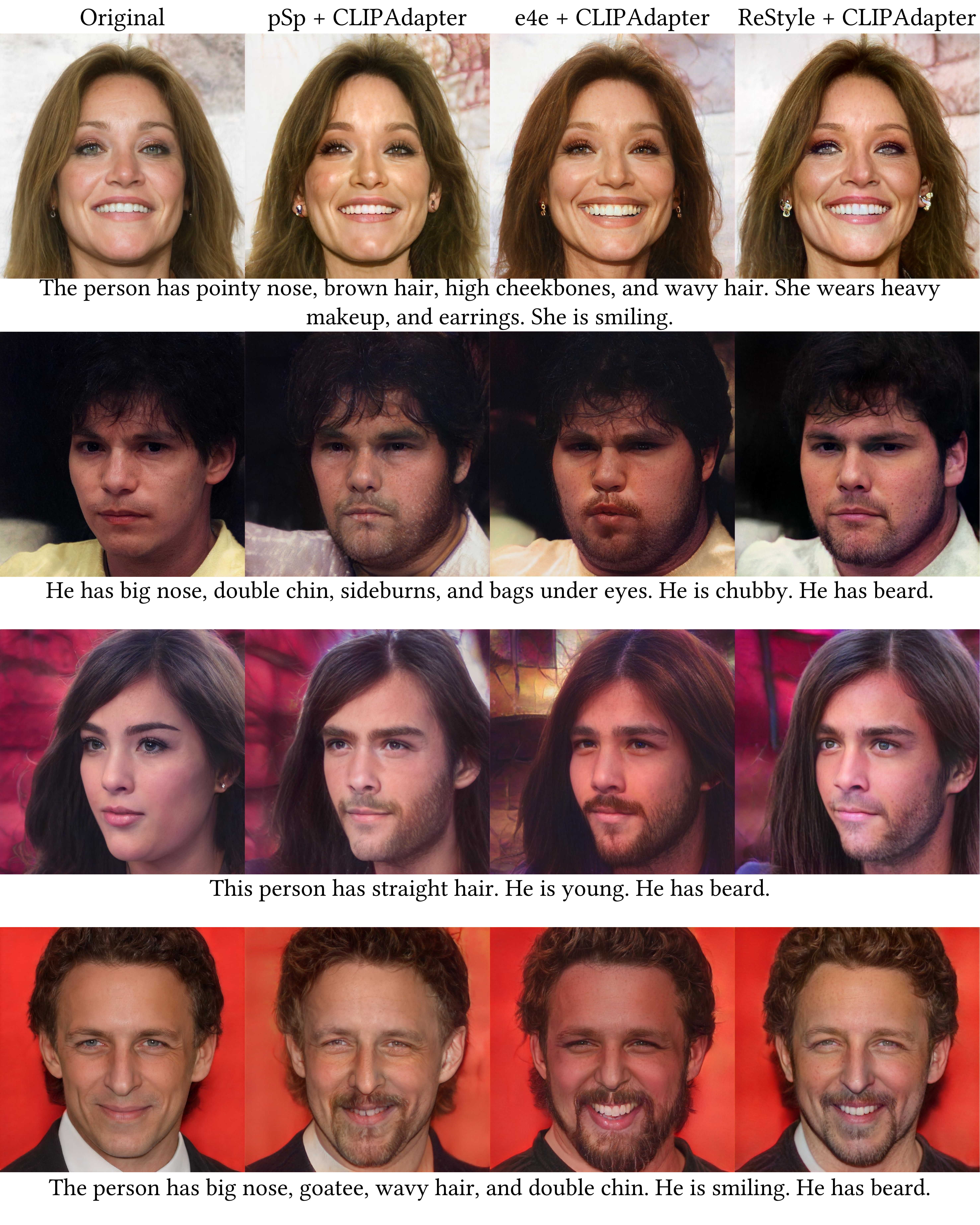}
  \caption{\textbf{Using different encoder networks with CLIPAdapter.} CLIPAdapter can be attached to various inversion networks and is able to find semantic directions in the latent space successfully, as demonstrated by the manipulations.
  }
  \label{fig:encodercomparison}
\end{figure}

CLIPAdapter is a light-weight adapter module that could be attached to any inversion network, as long as the encoder network processes some image feature maps during the inversion phase. CLIPAdapter modulates these feature maps using the CLIP embeddings of the description $t_{target}$ and we pass them through the rest of the encoder network. To support our claim, we trained CLIPInverter on different encoder networks, namely pSp~\cite{richardson2021encoding}, e4e~\cite{Tov2021} and ReStyle~\cite{alaluf2021restyle}. Table~\ref{tab:different_encoders_metrics} shows the quantitative comparisons between the pSp, e4e and ReStyle based models on the face images. All of the models achieve similar performance in terms of FID and CMP, but the e4e based model achieves superior AMA compared to the other models. Fig.~\ref{fig:encodercomparison} demonstrates the manipulation results obtained with different inversion networks equipped with the proposed CLIPAdapter module. Both models are able to edit images successfully. We observe that the e4e based model is able to achieve slightly more accurate outcomes. Therefore, we have used the e4e based network in our final model.

\begin{table}[h]
    \centering
    \caption{\textbf{Quantitative comparisons of CLIPAdapter used in different encoder networks.} The two models achieve similar scores, but e4e based model is superior in terms of manipulation accuracy. }
    \begin{tabular}{@{}l|ccc@{}}
        \toprule
        Encoder & FID $\downarrow$ & CMP $\uparrow$ & AMA $\uparrow$ \\
        \midrule
        pSp + CLIPAdapter & \textbf{93.483} & 0.221 & 54.143 \\
        e4e + CLIPAdapter & 97.210 & 0.221 & \textbf{61.429} \\
        ReStyle + CLIPAdapter & 96.197 & \textbf{0.222} & 50.000 \\
        \bottomrule
     \end{tabular}
    \label{tab:different_encoders_metrics}
\end{table}

\begin{figure*}[!t]
  \centering
  \includegraphics[width=\linewidth]{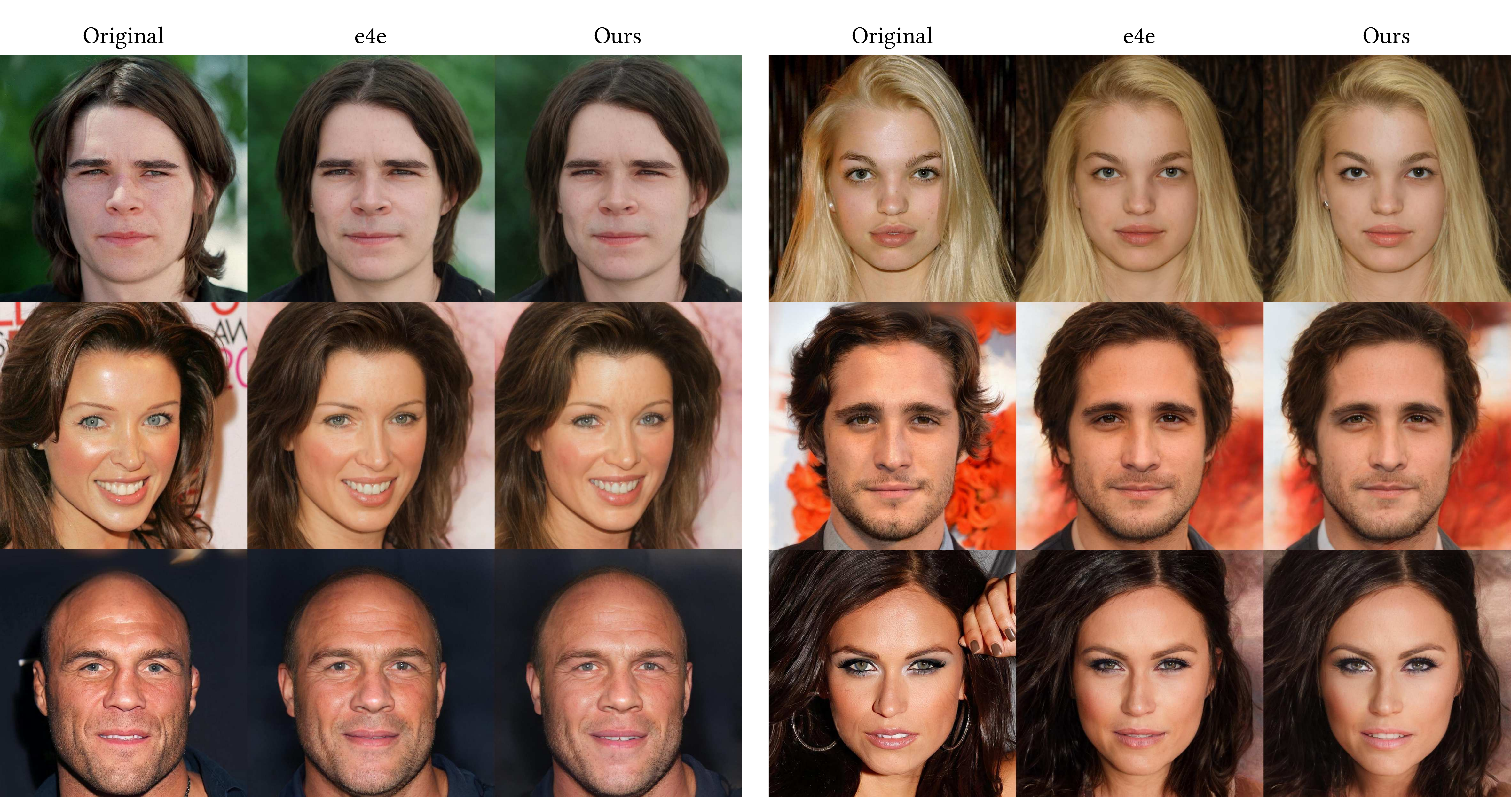}
  \caption{\textbf{Using CLIP image embeddings in CLIPAdapter to modulate the feature maps.} Our approach is able to achieve inversions that are more faithful to the inputs compared to e4e, such as the earrings in the first row or the slightly open mouth in the third row.
  }
  \label{fig:imagecondinversion}
\end{figure*}

\section{CLIPAdapter for Inversion}
\label{sec:imagecondinversion}
In our main paper, we demonstrate that CLIPAdapter finds more accurate semantic directions in the latent space for editing a provided input image. We also examine whether similar feature modulation scheme improves the GAN inversion for a more accurate image reconstruction. In this setup, we used CLIPAdapter in a pre-trained e4e~\cite{Tov2021} encoder. Instead of the CLIP text embeddings of a target textual description $t_{target}$, we used the CLIP image embedding of the input image $x_{in}$ to modulate the image feature maps during the inversion process. We use a pre-trained and frozen StyleGAN2 generator and we do not employ CLIPRemapper for this experiment. Fig.~\ref{fig:imagecondinversion} shows some comparisons between the e4e inversions and inversions using our setup described above. Our approach slightly improves upon the e4e inversions. For instance, the earrings are more accurately represented in our inversions in the first row, as the e4e inversions add an earring to the first image or remove it from the second image. We see from the second row that our approach represents rosy cheeks or the jacket/necktie better than the e4e model. Moreover, the third row shows that the gap between the lips are much more accurately preserved with our approach. 

Quantitative comparisons between e4e and our approach is also given in Table~\ref{tab:imagecondinversion_quantitative}. We use the L2, LPIPS, ID similarity, FID, KID, PSNR and MS-SSIM metrics for these comparisons. As most of the metrics suggest, our approach improves upon the inversion performance of e4e, supporting our claim that our CLIP-guided adapter can lead to more meaningful latent codes and thus more accurate reconstructions.

\begin{table}[h]
    \centering
    \caption{\textbf{Quantitative comparisons against e4e for inversion with CLIPAdapter.} As most of the metrics confirm, our CLIPAdapter framework improves upon the performance of the e4e network.}
    \resizebox{\linewidth}{!}{
    \begin{tabular}{lc@{$\;\;\,$}c@{$\;\;\,$}c@{$\;\;\,$}c@{\;\;\,}c@{\;\;\,}c@{\;\;\,}c}
        \toprule
        Model & L2 $\downarrow$ & LPIPS $\downarrow$ & ID $\uparrow$ & FID $\downarrow$ & KID $\downarrow$ & PSNR $\uparrow$ & MS-SSIM $\uparrow$\\
        \midrule
        e4e & 0.047 & \textbf{0.198} & \textbf{0.493} & 36.120 & 13.288 & 19.119 & 0.619 \\
        Ours & \textbf{0.046} & 0.199 & 0.487 & \textbf{33.826} & \textbf{10.860} & \textbf{19.194} & \textbf{0.622} \\
        \bottomrule
     \end{tabular}}
    \label{tab:imagecondinversion_quantitative}
\end{table}

\section{Additional Qualitative Results}
\label{sec:more_results}
\begin{figure*}
\begin{tabular}{cc}
\subfloat[This man has bags under eyes. He is chubby. He has beard.]{\includegraphics[width=0.5\linewidth]{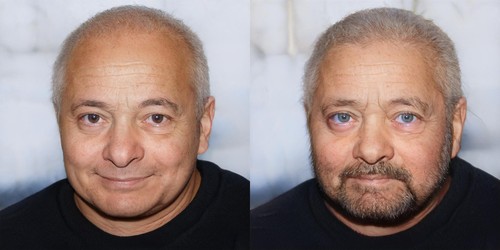}} &
\subfloat[This smiling woman has bangs.]{\includegraphics[width=0.5\linewidth]{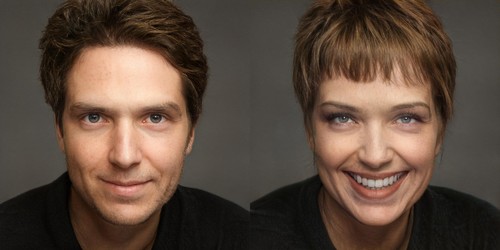}}\\
\subfloat[This man has wavy hair and wears necktie. He is smiling, and young.]
{\includegraphics[width=0.5\linewidth]{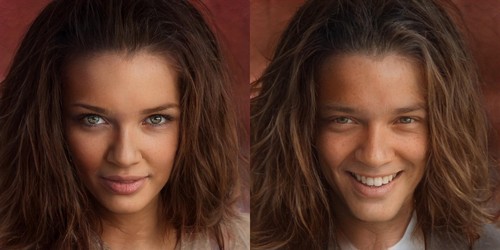}} &
\subfloat[This woman is wearing lipstick. She has eyeglasses. She is smiling.]{\includegraphics[width=0.5\linewidth]{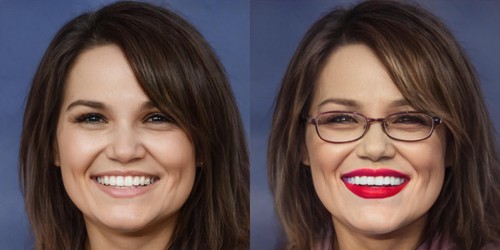}}\\
\subfloat[The man has mustache, bags under eyes, big nose, goatee, and straight hair.]{\includegraphics[width=0.5\linewidth]{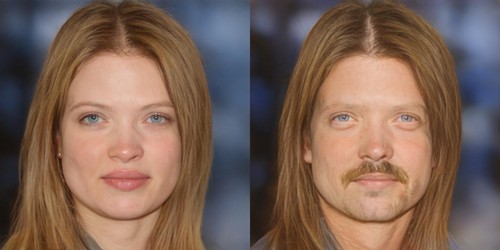}} &
\subfloat[This man is wearing necktie. He has high cheekbones. He is smiling.]{\includegraphics[width=0.5\linewidth]{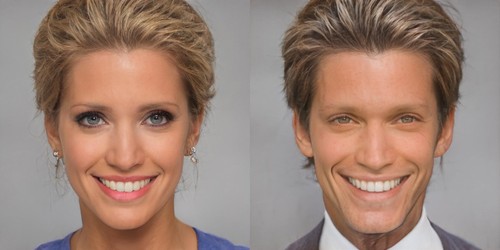}}\\
\subfloat[The man has big nose. He is smiling, and chubby. He has beard.]{\includegraphics[width=0.5\linewidth]{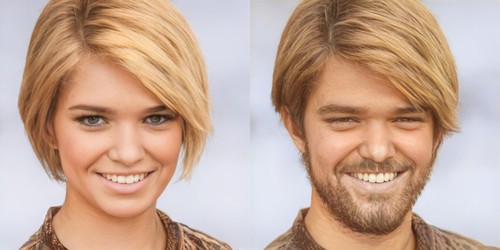}} &
\subfloat[The woman is smiling and has bangs, brown hair, and mouth slightly open.]{\includegraphics[width=0.5\linewidth]{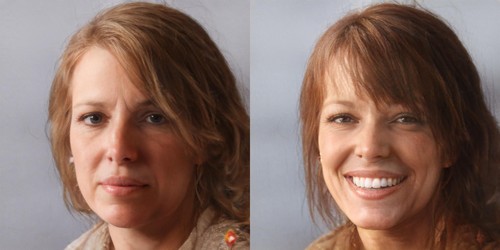}}
\end{tabular}
\caption{Additional text-driven manipulation results on the human face images obtained with our CLIPInverter approach.}
\label{fig:qualitative_faces}
\end{figure*}

\begin{figure*}
\begin{tabular}{cc}
\subfloat[A calico cat with cinnamon hair.]{\includegraphics[width=0.5\linewidth]{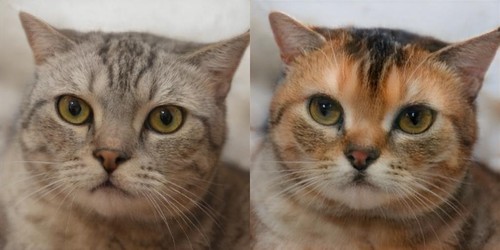}} &
\subfloat[A cat with grey hair.]{\includegraphics[width=0.5\linewidth]{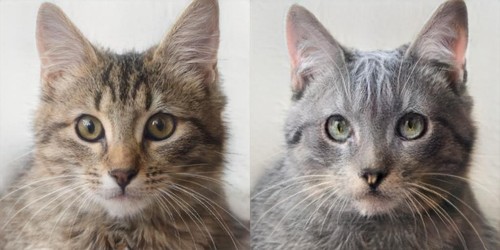}}\\
\subfloat[A grumpy cat. ]{\includegraphics[width=0.5\linewidth]{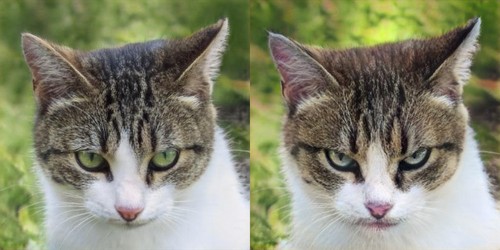}} &
\subfloat[A ragdoll kitty.]{\includegraphics[width=0.5\linewidth]{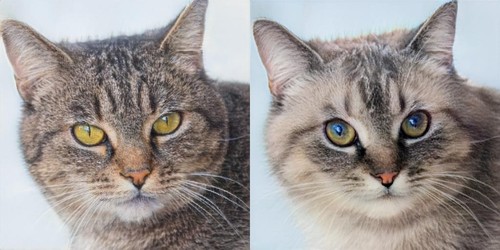}}\\
\subfloat[A fearful cat with grey hair.]{\includegraphics[width=0.5\linewidth]{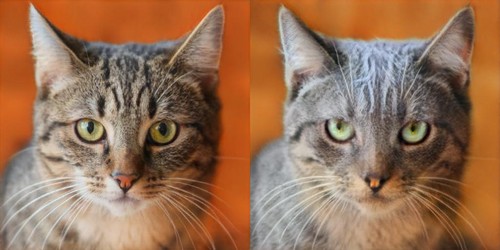}} &
\subfloat[A kitty with ginger hair.]{\includegraphics[width=0.5\linewidth]{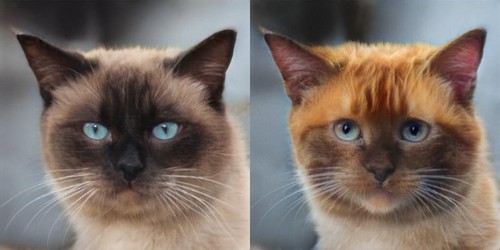}}\\
\subfloat[An elderly cat.]{\includegraphics[width=0.5\linewidth]{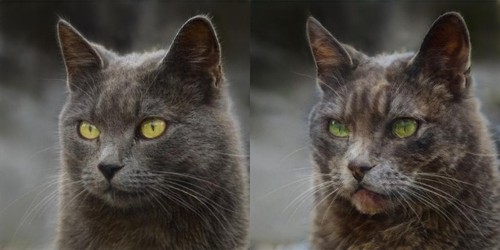}} &
\subfloat[A calico kitty.]{\includegraphics[width=0.5\linewidth]{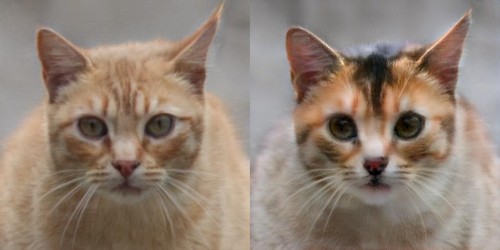}}

\end{tabular}
\caption{Additional text-driven manipulation results on the cat images obtained with our CLIPInverter approach.}
\label{fig:qualitative_cats}
\end{figure*}

\begin{figure*}
\begin{tabular}{cc}
\subfloat[This small bird has a reddish brown crown and white belly. its wingbars are white while its primaries, secondaries, and coverts are light brown and
black.]{\includegraphics[width=0.5\linewidth]{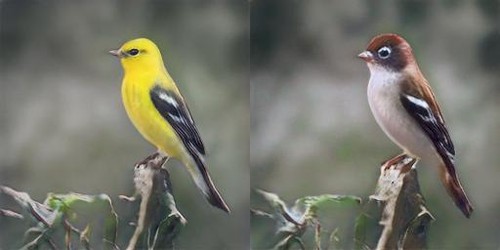}} &
\subfloat[This gray bird has a black cheek patch, black primaries and a black tail, while its throat is white.]{\includegraphics[width=0.5\linewidth]{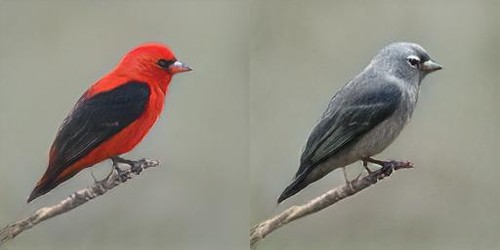}}\\
\subfloat[This bird is yellow with grey and has a very short beak. ]{\includegraphics[width=0.5\linewidth]{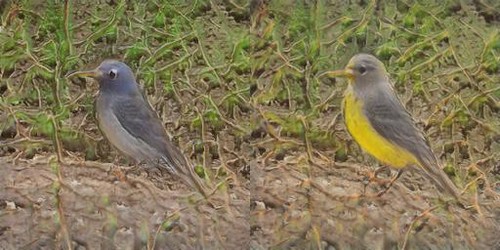}} &
\subfloat[This bird has wings that are brown and has a yellow belly.]{\includegraphics[width=0.5\linewidth]{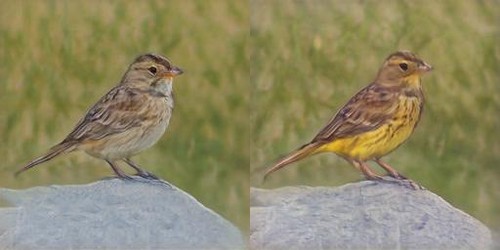}}\\
\subfloat[This tiny colorful bird has a yellow belly and short beak.]{\includegraphics[width=0.5\linewidth]{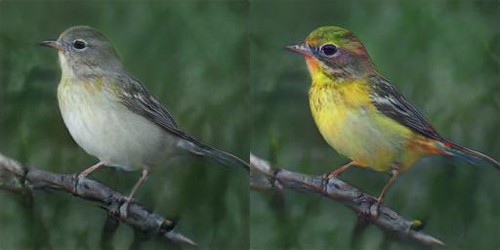}} &
\subfloat[This bird has a white belly and breast with a short pointy bill]{\includegraphics[width=0.5\linewidth]{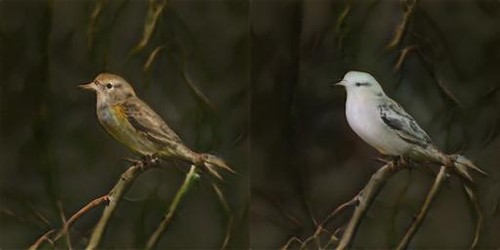}}\\
\subfloat[This bird has wings that are green and has a yellow belly.]{\includegraphics[width=0.5\linewidth]{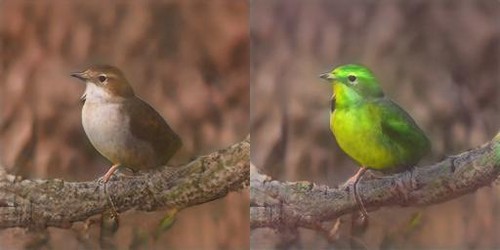}} &
\subfloat[Tiny grey and white bird with black eyes and a sharp beak]{\includegraphics[width=0.5\linewidth]{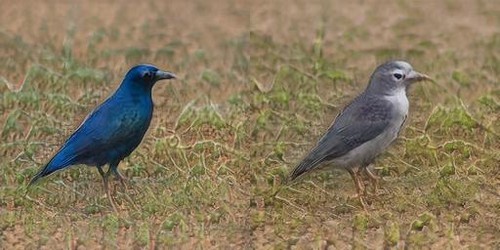}}

\end{tabular}
\vspace{-0.25cm}
\caption{Additional text-driven manipulation results on the bird images obtained with our CLIPInverter approach.}
\label{fig:qualitative_birds}
\end{figure*}

\begin{figure*}[!t]
  \centering
  \includegraphics[width=\linewidth]{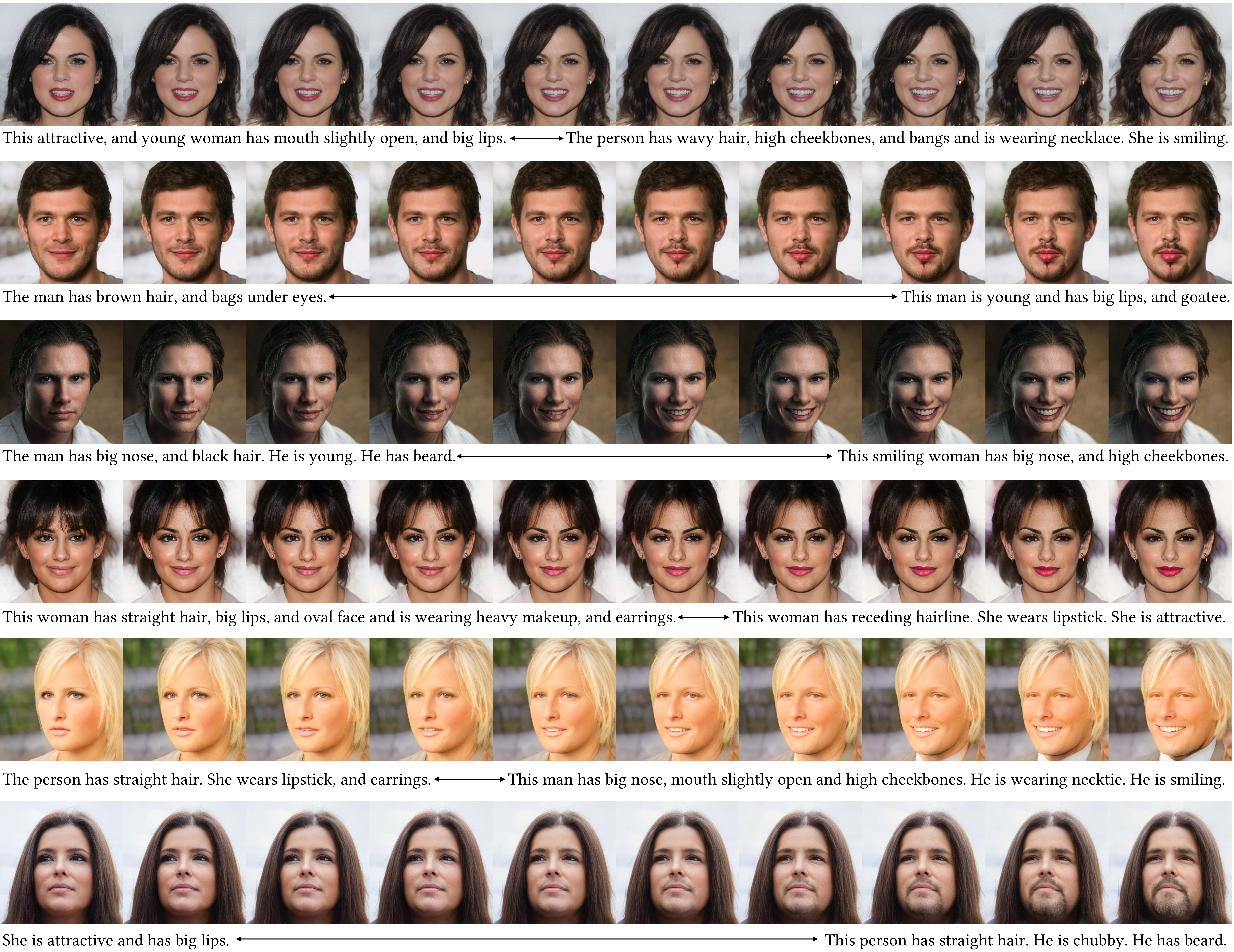}
  \caption{Additional continuous manipulation results for some human face images. For reference, we provide the original (\textit{left}) and the target descriptions (\textit{right}) below each row.}
  \label{fig:faces_interp}
\end{figure*}

\begin{figure*}[!t]
  \centering
  \includegraphics[width=\linewidth]{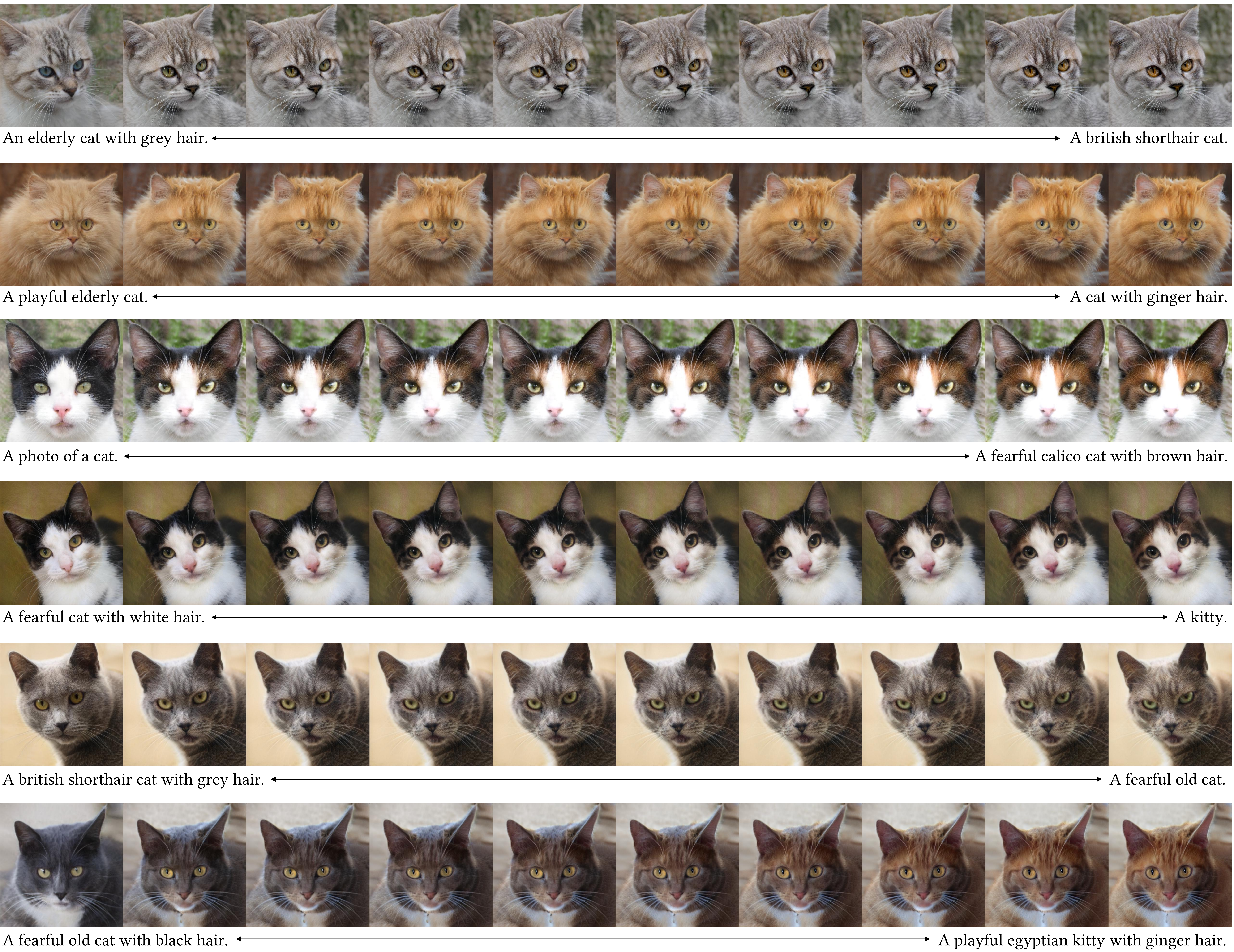}
  \caption{Additional continuous manipulation results for some cat images. For reference, we provide the original (\textit{left}) and the target descriptions (\textit{right}) below each row.}
  \label{fig:cats_interp}
\end{figure*}

\begin{figure*}[!t]
  \centering
  \includegraphics[width=\linewidth]{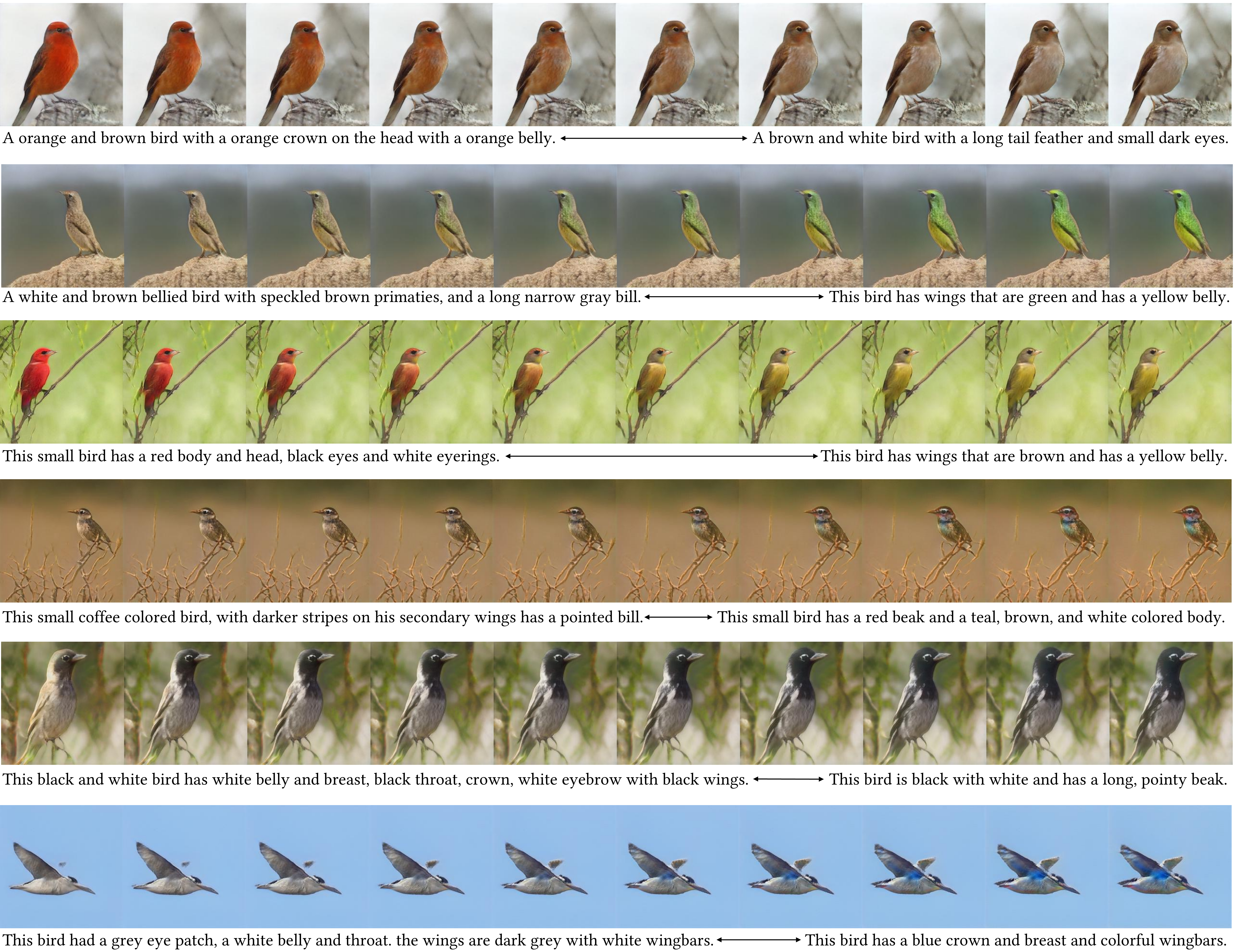}
  \caption{Additional continuous manipulation results for some bird images. For reference, we provide the original (\textit{left}) and the target descriptions (\textit{right}) below each row.}
  \label{fig:birds_interp}
\end{figure*}

\begin{figure*}
\begin{tabular}{cc}
\subfloat[]{\includegraphics[width=0.5\linewidth]{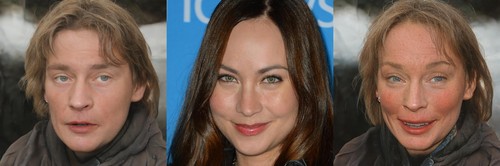}}&
\subfloat[]{\includegraphics[width=0.5\linewidth]{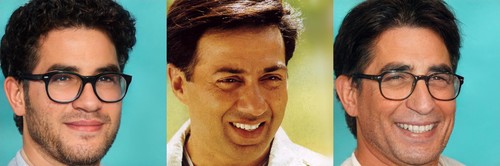}}\\
\subfloat[]{\includegraphics[width=0.5\linewidth]{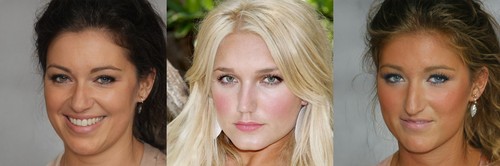}}&
\subfloat[]{\includegraphics[width=0.5\linewidth]{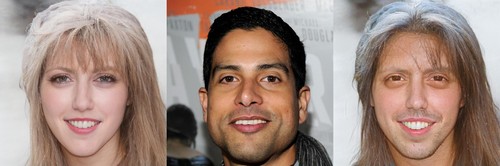}}\\
\subfloat[]{\includegraphics[width=0.5\linewidth]{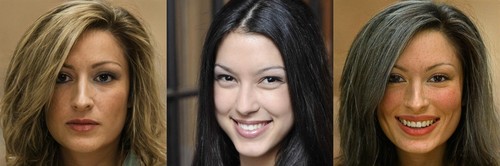}}&
\subfloat[]{\includegraphics[width=0.5\linewidth]{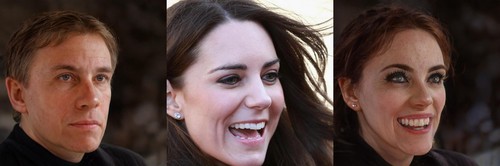}}\\
\subfloat[]{\includegraphics[width=0.5\linewidth]{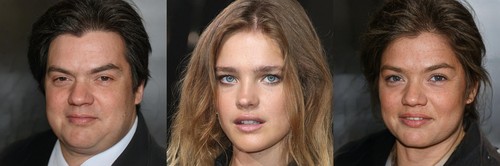}}&
\subfloat[]{\includegraphics[width=0.5\linewidth]{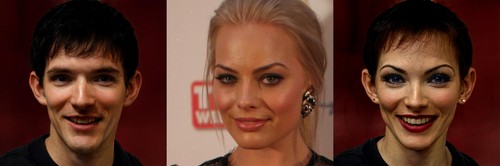}}

\end{tabular}
\caption{Additional image-guided manipulation results on the human face images, showing the original image (\textit{left}), the condition image (\textit{middle}), and the manipulated image (\textit{right}).}
\label{fig:faces_cond}
\end{figure*}

\begin{figure*}
\begin{tabular}{cc}
\subfloat[]{\includegraphics[width=0.5\linewidth]{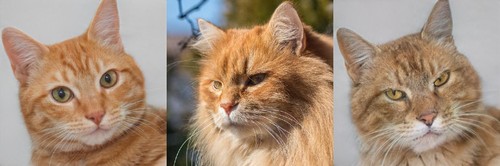}}&
\subfloat[]{\includegraphics[width=0.5\linewidth]{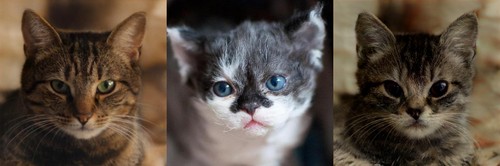}}\\
\subfloat[]{\includegraphics[width=0.5\linewidth]{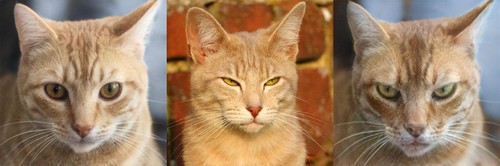}}&
\subfloat[]{\includegraphics[width=0.5\linewidth]{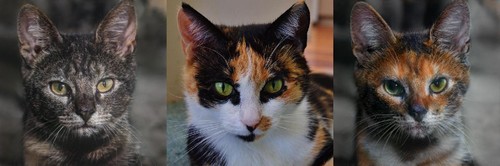}}\\
\subfloat[]{\includegraphics[width=0.5\linewidth]{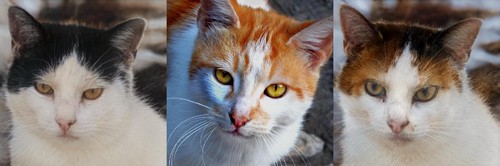}}&
\subfloat[]{\includegraphics[width=0.5\linewidth]{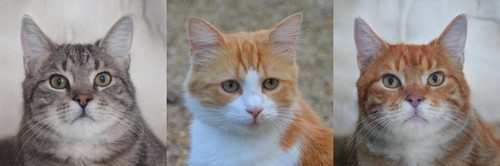}}\\
\subfloat[]{\includegraphics[width=0.5\linewidth]{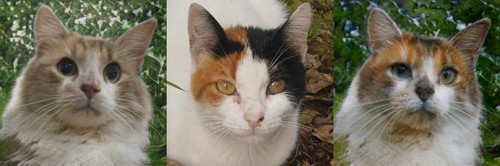}}&
\subfloat[]{\includegraphics[width=0.5\linewidth]{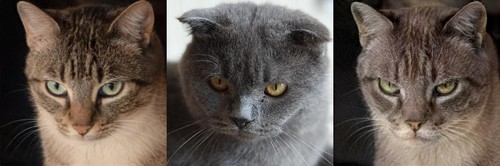}}

\end{tabular}
\caption{Additional image-guided manipulation results on the cat images, showing the original image (\textit{left}), the condition image (\textit{middle}), and the manipulated image (\textit{right}).}
\label{fig:cats_cond}
\end{figure*}

\begin{figure*}
\begin{tabular}{cc}
\subfloat[]{\includegraphics[width=0.5\linewidth]{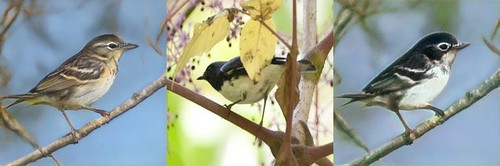}}&
\subfloat[]{\includegraphics[width=0.5\linewidth]{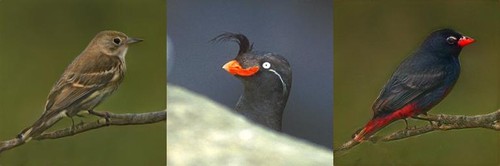}}\\
\subfloat[]{\includegraphics[width=0.5\linewidth]{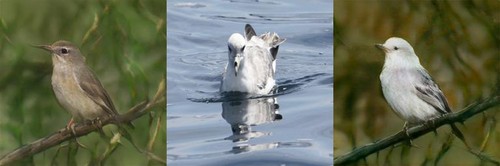}}&
\subfloat[]{\includegraphics[width=0.5\linewidth]{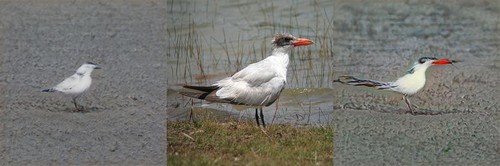}}\\
\subfloat[]{\includegraphics[width=0.5\linewidth]{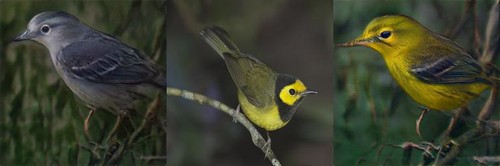}}&
\subfloat[]{\includegraphics[width=0.5\linewidth]{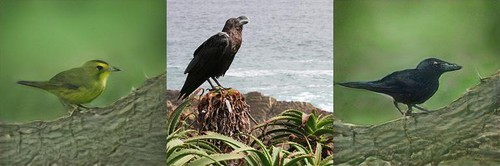}}\\
\subfloat[]{\includegraphics[width=0.5\linewidth]{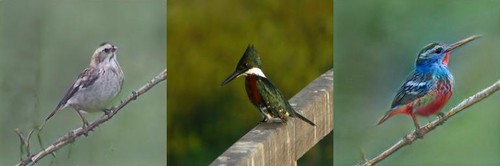}}&
\subfloat[]{\includegraphics[width=0.5\linewidth]{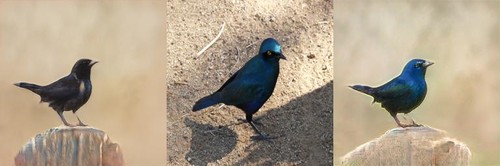}}

\end{tabular}
\caption{Additional image-guided manipulation results on the bird images, showing the original image (\textit{left}), the condition image (\textit{middle}), and the manipulated image (\textit{right}).}
\label{fig:birds_cond}
\end{figure*}

In Fig.~\ref{fig:qualitative_faces}-\ref{fig:qualitative_birds}, we provide additional text-driven editing results of our proposed model on human faces, cats and birds images, respectively. Additionally, in Fig.~\ref{fig:faces_interp}-\ref{fig:birds_interp}, we provide continuous manipulations obtained by our method by interpolating the predicted residual latent code for a number of images and some target descriptions. Finally, in Fig.~\ref{fig:faces_cond}-\ref{fig:birds_cond}, we present additional image-conditioned manipulation results, again considering images of human faces, cats and birds, respectively.

\captionsetup[subfloat]{labelformat=empty}

\section{Additional Performance Comparisons}
\label{sec:more_comparisons}
\begin{figure*}[!t]
  \centering
  \includegraphics[width=0.9\linewidth]{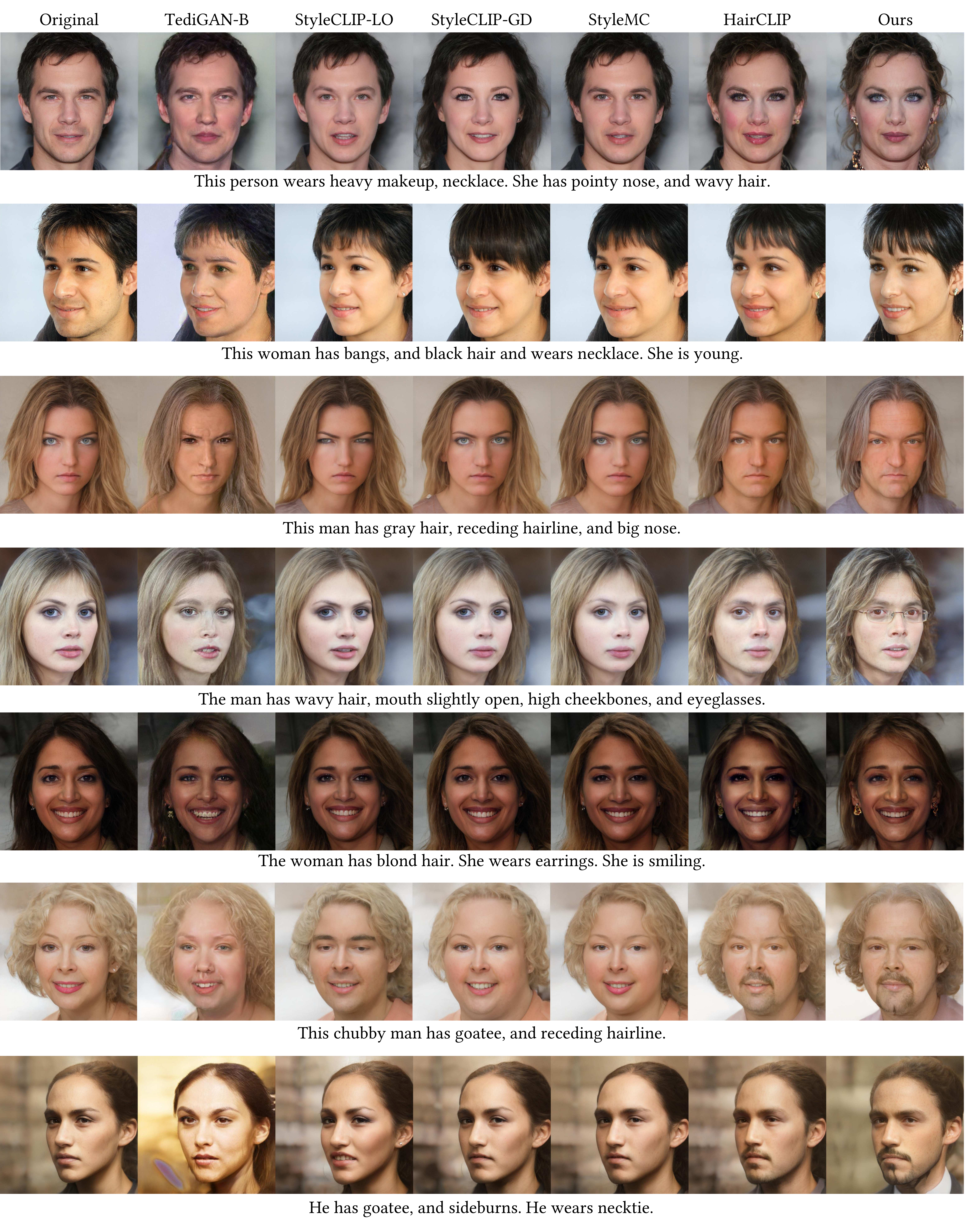}
  \caption{Additional comparisons against the state-of-the-art text-driven manipulation methods on the human face images. Our method performs edits relevant to the given target descriptions more accurately than the competing approaches.}
  \label{fig:more_comparisons}
\end{figure*}

\begin{figure*}[!t]
  \centering
  \includegraphics[width=0.8\linewidth]{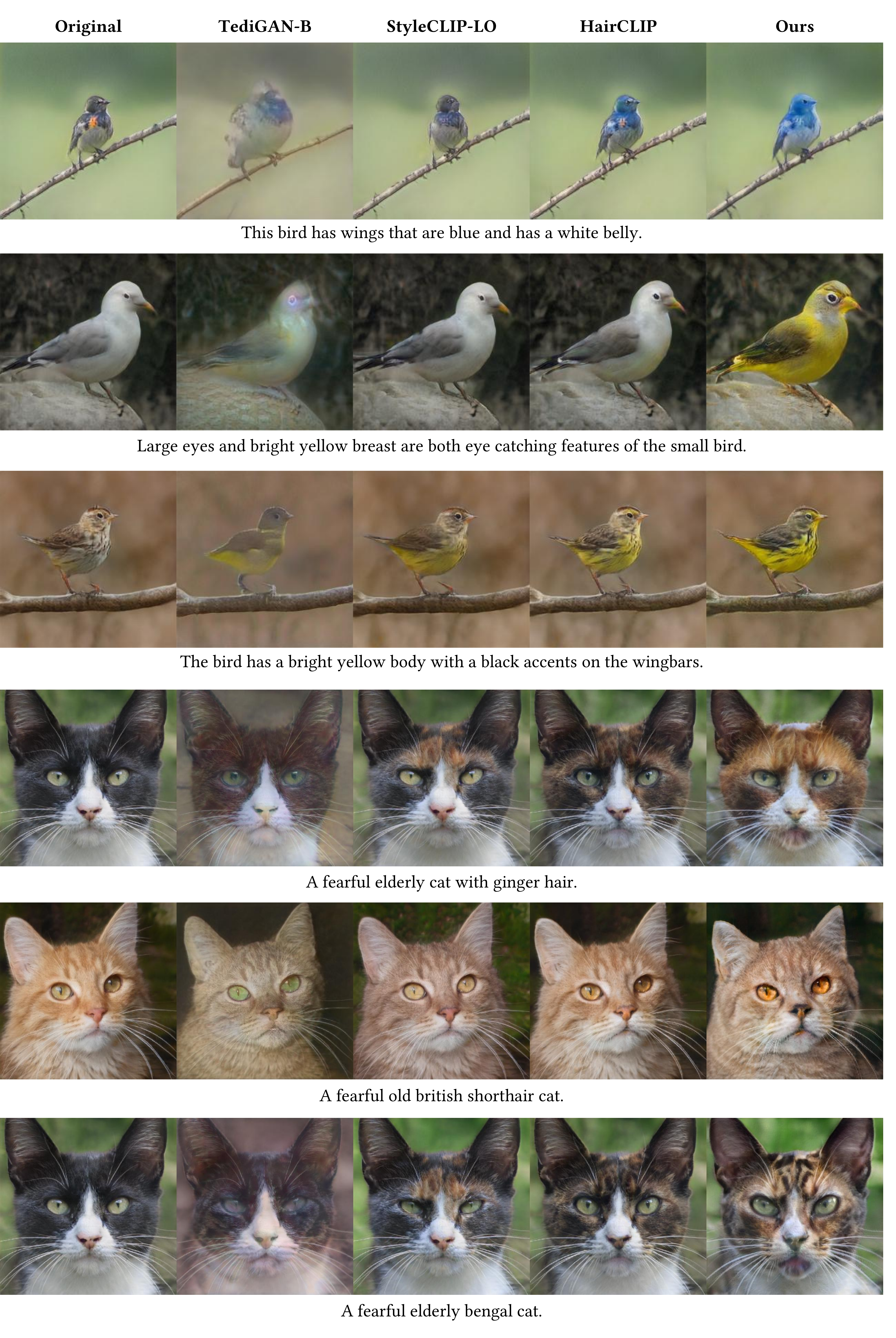}
  \vspace{-0.5cm}
  \caption{Additional comparisons against the state-of-the-art text-driven manipulation methods on the bird and the cat images. Our method performs edits relevant to the given target descriptions more accurately than the competing approaches.}
  \label{fig:more_comparisons_birdcat}
\end{figure*}

In Fig.~\ref{fig:more_comparisons}, we present additional comparisons of our CLIPInverter model against the TediGAN~\cite{xia2021tedigan}, the StyleCLIP~\cite{Patashnik_2021_ICCV}, the StyleMC~\cite{kocasari2021} and the HairCLIP~\cite{wei2022hairclip} models on the CelebA face dataset. In Fig.~\ref{fig:more_comparisons_birdcat}, we present additional comparisons of our CLIPInverter against the competing approach on the CUB bird and AFHQ-Cats datasets.

\section{Comparisons Against Diffusion-Based Approaches}
\label{sec:diffusion_comparison}
\begin{figure*}[!t]
  \centering
  \includegraphics[width=0.65\linewidth]{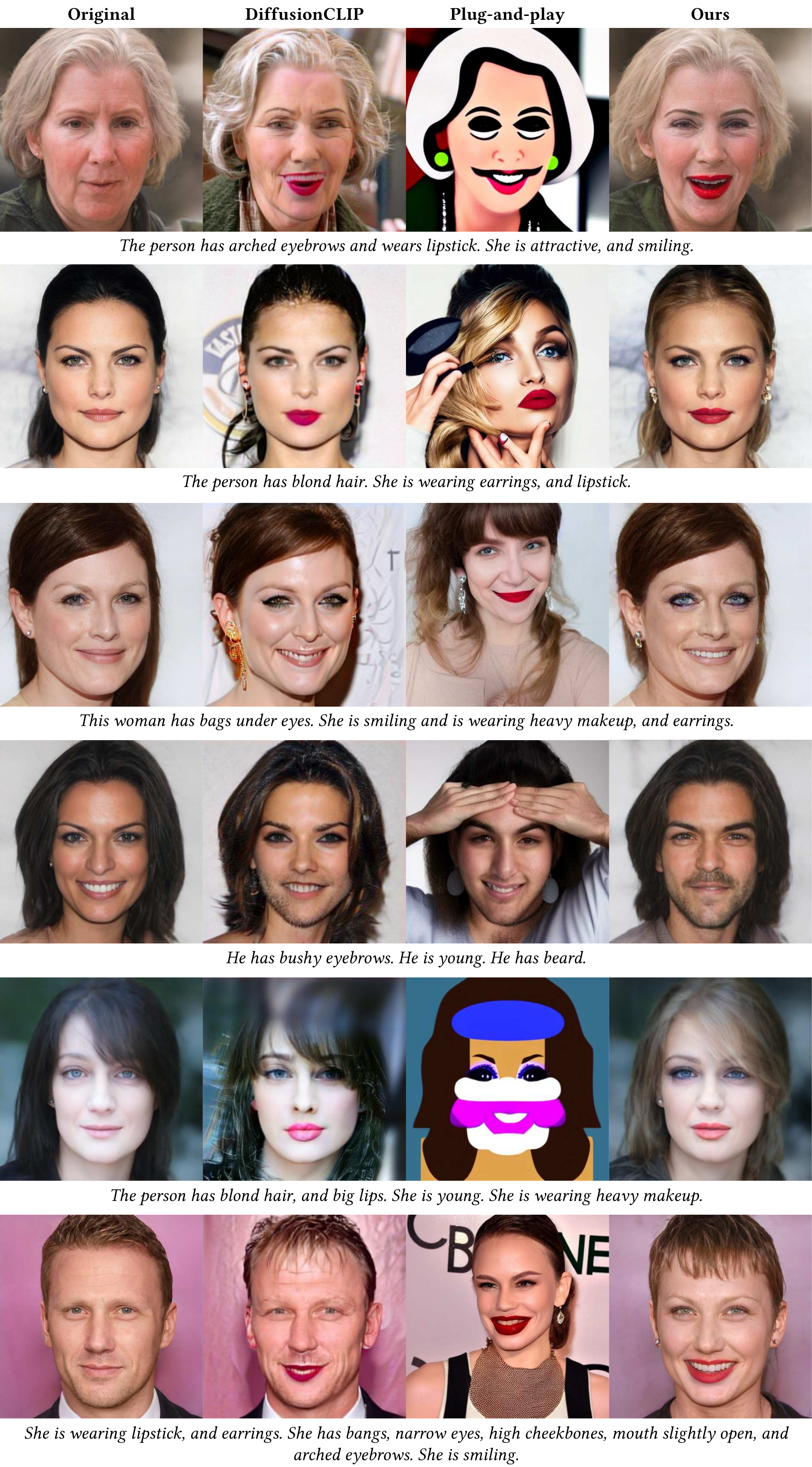}
  \caption{Qualitative comparions against diffusion-based text-guided editing methods. Our method performs edits relevant to the given target descriptions more accurately and generates realistic looking images.}
  \label{fig:diffusion_qualitative}
\end{figure*}

We compare our approach against the DiffiusionCLIP~\cite{Kim_2022_CVPR} and Plug-and-Play~\cite{pnpDiffusion2022} approaches qualitatively and quantitatively. Table~\ref{tab:diffusion_quantitative} shows the quantitative comparisons and Fig.~\ref{fig:diffusion_qualitative} shows qualitative comparisons against these methods on the CelebA~\cite{liu2015faceattributes} dataset. We observe that the diffusion-based models achieve good FID scores thanks to their strong synthesis capabilities. However, our approach is able to manipulate the images with much higher accuracy.  DiffusionCLIP struggles to apply all the manipulations and yields results with some artifacts, whereas Plug-and-Play outputs cartoonish and unrealistic looking results.

\begin{table}[!h]
    \caption{\textbf{Quantitative comparisons on the CelebA dataset against diffusion-based models.} }
    \resizebox{\linewidth}{!}{  
\begin{tabular}{l|cccc}
        \toprule
             & \multicolumn{1}{c}{FID~$\downarrow$} & \multicolumn{1}{c}{CMP~$\uparrow$} & \multicolumn{1}{c}{AMA (Single)~$\uparrow$} & \multicolumn{1}{c}{AMA (Multiple)~$\uparrow$} \\
             
        \midrule
DiffusionCLIP & \textbf{29.280} & \textbf{0.243} & 26.000 & 4.857 \\
Plug-and-Play & \underline{68.287} & 0.199 & \underline{27.429} & \underline{7.143}  \\
Ours         & 97.210 & \underline{0.221} & \textbf{61.429} & \textbf{41.714} \\
         \bottomrule
\end{tabular}
}
\label{tab:diffusion_quantitative}
\end{table}

\section{Comparisons Against HairCLIP on Unseen Text Prompts}
\label{sec:unseen_comparison}
\begin{figure*}[!t]
  \centering
  \includegraphics[width=\linewidth]{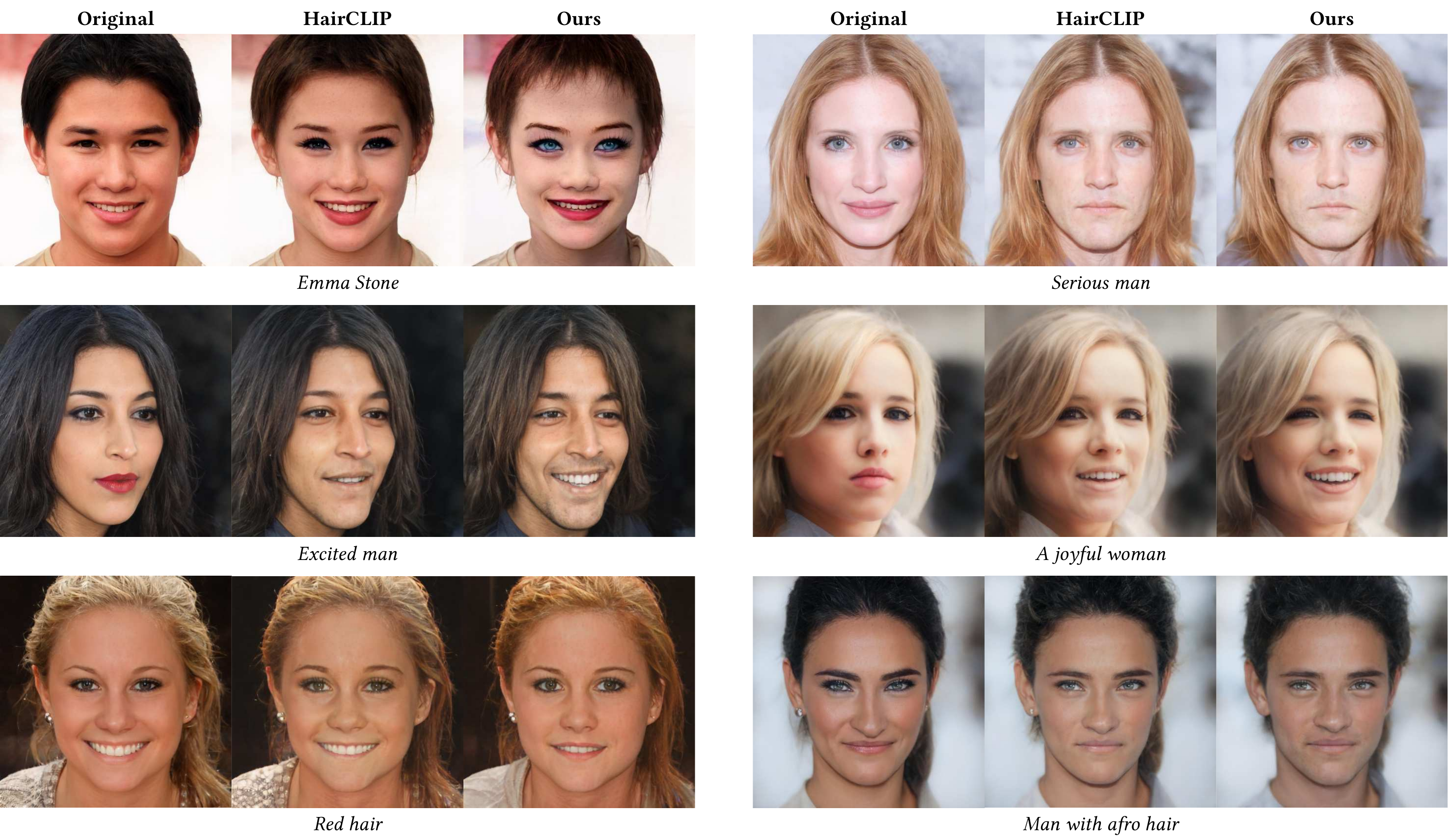}
  \caption{\textbf{Manipulation results of HairCLIP and CLIPInverter with descriptions that are unseen during training.} Our model learns a semantic alignment with the CLIP space and is able to apply the manipulations even when the descriptions are not included in the training set.}
  \label{fig:unseen_results}
\end{figure*}

Fig.~\ref{fig:unseen_results} showcases manipulation results achieved by CLIPInverter and HairCLIP using textual descriptions that were not encountered during the training of either model. Both models demonstrate remarkable editing ability as they establish a semantic alignment with the CLIP space, enabling them to generate plausible results even when presented with unseen target descriptions. However, it is worth noting that the models perform significantly better when the descriptions are semantically similar to those in the training set. This discrepancy becomes apparent in the last example of the figure, where the descriptions deviate a lot from the training distribution. To quantitative evaluate the performance change, we consider novel compositions that do not appear in the training set, which are listed below. Table~\ref{tab:unseen} shows the results. Overall, our CLIPInverter exhibits a superior ability to align with the CLIP space and applies manipulations with higher accuracy compared to HairCLIP. Its performance on unseen text prompt is even higher than the performance of HairCLIP on seen text prompts.

\begin{multicols}{2}
\begin{itemize}
    \item {\textit{heavy makeup + mustache}}
    \item {\textit{chubby + wearing lipstick}}
    \item {\textit{blond hair + mustache}}
    \item {\textit{wearing earrings + wearing necktie}}
    \item {\textit{eyeglasses + smiling + wearing earrings}}
    \item {\textit{mouth slightly open + wearing earrings + wearing lipstick}}
    \item {\textit{bangs + goatee + gray hair}}      
\end{itemize}
\end{multicols}
\begin{table}[!h]
\caption{\textbf{Quantitative comparisons against HairCLIP on unseen text prompts}. On the CelebA dataset, we have performed experiments to evaluate the generalization capabilities of the methods on text prompts not encountered during training. For the sake of completeness, we also include the performances with previously seen prompts.}
    \centering
    \begin{tabular}{l|ccc|ccc}
    \toprule
         & \multicolumn{3}{c|}{Seen Text Prompts} & \multicolumn{3}{c}{Unseen Text Prompts}\\
                  & CMP $\uparrow$ & AMA $\uparrow$& ID $\uparrow$ & CMP $\uparrow$ & AMA $\uparrow$ & ID$\uparrow$\\ \midrule
         HairCLIP &  0.221 & 15.143 & 55.25 & 0.229 & 6.561 & 53.55\\
         Ours & 0.224 & 41.714 & 46.12 & 0.236 & 16.286 & 47.49\\
    \bottomrule
    \end{tabular}
    
    \label{tab:unseen}
\end{table}

\section{Limitations due to StyleGAN Inversion}
\label{sec:supplementary_limitations}
\begin{figure*}[!t]
  \centering
  \includegraphics[width=\linewidth]{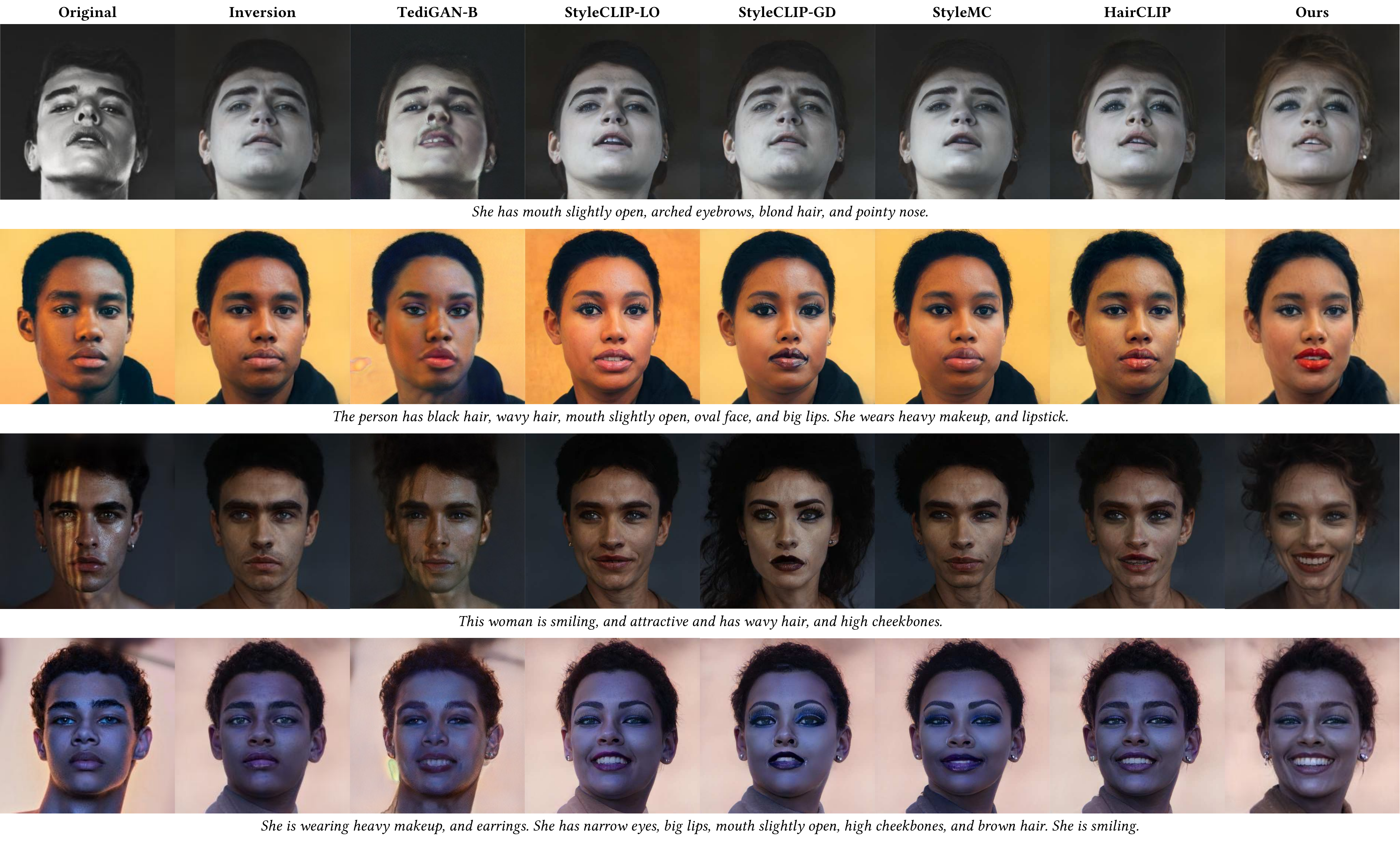}
  \caption{Qualitative comparisons of manipulations with out of distribution input images.}
  \label{fig:supp_limitations}
\end{figure*}

The common practice to edit images using pre-trained StyleGAN models is to first invert the inputs to the latent space using a pre-trained GAN inversion method. However, when the input images exhibit unconventional poses, shadows, and other complex characteristics, the inversion model encounters difficulties in preserving all the intricate details. As a consequence, the manipulation results often deviate from the original inputs. In Fig.~\ref{fig:supp_limitations}, we present results for the competing approach as well as our CLIPInverter for these challenging cases.

\section{Analysis of CLIPRemapper }
\label{sec:clipremapper}

\begin{figure*}[!t]
  \centering
  \includegraphics[width=0.7\linewidth]{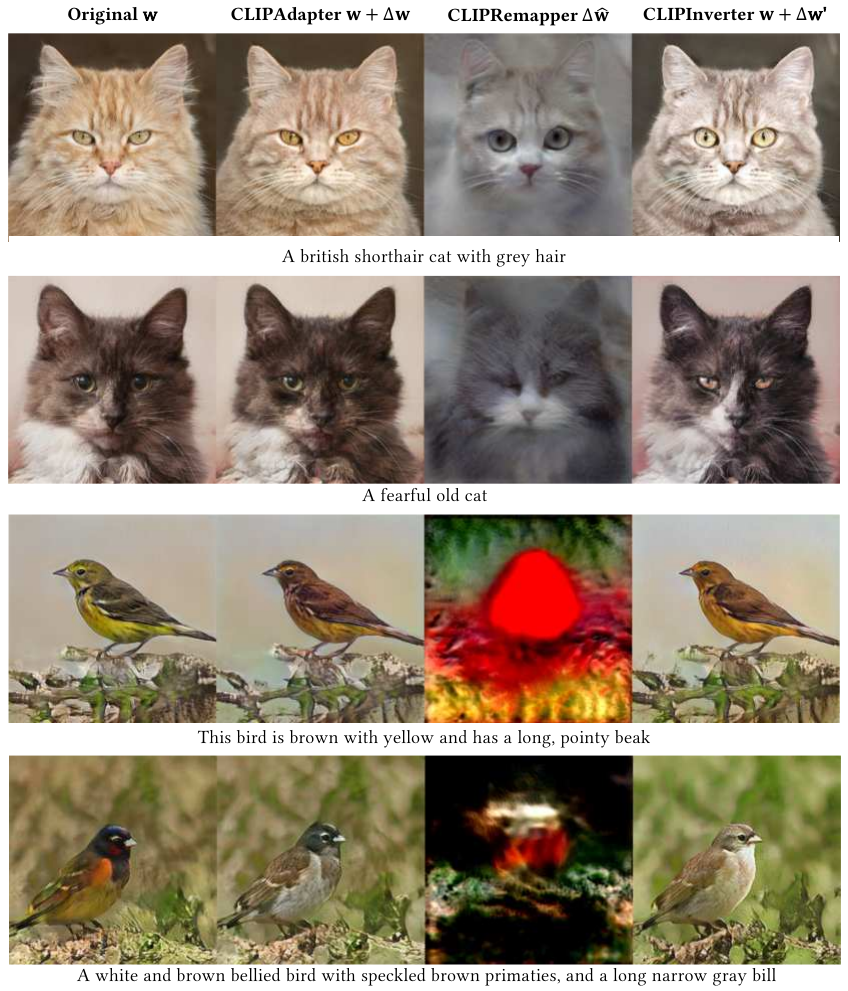}
  \caption{\textbf{Visualization of the latent code correction operation via CLIPRemapper on cats and birds images.} For each sample manipulation operation, we show the initial editing results generated solely by CLIPAdapter, the images generated via CLIPRemapper, and the final manipulations by CLIPInverter obtained by the suggested correction scheme. Our refinement module improves the quality of the intended manipulations on both domains.} 
  \label{fig:clipremapper}
\end{figure*}

Fig.~\ref{fig:clipremapper} visually depicts the refinement process conducted by the CLIPRemapper module using sample images from the cat and bird domains. It effectively demonstrates how this proposed scheme enhances the quality of image manipulations based on the provided target descriptions. Notably, the refinements introduced by the CLIPRemapper module are particularly prominent in the cat domain, which exhibits a higher degree of structural complexity compared to the bird domain. However, it is worth noting that the CLIPRemapper output for birds should be interpreted with caution. Since the output corresponding to the zero latent code lies outside the distribution, unlike the human and cat domains, any latent code in its vicinity, including the output of CLIPRemapper $\Delta \widehat{w}$, is expected to produce an out-of-distribution image. Additionally, the CLIPRemapper output may not exhibit all the characteristics mentioned in the text prompt, as the module is trained subsequent to the CLIPAdapter and primarily focuses on improving cases where the CLIPAdapter may fall short.

\clearpage
\bibliographystyle{ACM-Reference-Format}
\bibliography{bibliography.bib}